\documentclass{article}
\newif\ifchinese
\chinesefalse
\PassOptionsToPackage{authoryear,round}{natbib}
\usepackage{iftex}
\usepackage[preprint]{neurips_2026}
\ifPDFTeX
  \usepackage{lmodern}
  \usepackage[T1]{fontenc}
  \usepackage[utf8]{inputenc}
  \ifchinese
    \usepackage[UTF8,fontset=none]{ctex}

    \AtBeginDocument{\raggedright}
  \fi
\else
  \ifchinese
    \usepackage[UTF8,fontset=fandol]{ctex}
  \fi
  \usepackage{fontspec}
\fi
\usepackage{amsmath,amssymb,booktabs,graphicx,xcolor}
\usepackage{placeins} 
\usepackage{tikz}
\usetikzlibrary{arrows.meta,positioning}
\usepackage{hyperref}
\hypersetup{colorlinks=true,linkcolor=blue!45!black,citecolor=blue!45!black,urlcolor=blue!45!black}
\newcommand{\bi}[2]{\ifchinese#1\else#2\fi}
\newcommand{\papertablestyle}{\footnotesize\setlength{\tabcolsep}{2pt}\renewcommand{\arraystretch}{1.16}}
\newcommand{\thd}[1]{#1}
\newcommand{\tgroup}[1]{\textbf{\boldmath#1}}
\newcommand{\thunit}[2]{\shortstack{#1\\#2}}
\newcommand{\tnum}[2]{\begingroup\settowidth{\dimen0}{#1}\makebox[\dimen0][r]{#2}\endgroup}
\newcommand{\tablenote}[1]{\par\smallskip\begin{minipage}{\linewidth}\footnotesize\raggedright#1\end{minipage}}

\newcommand{\qscore}{Q_{\mathrm{mean}}}
\definecolor{novablue}{HTML}{176B87}
\definecolor{generalorange}{HTML}{B96924}
\definecolor{filegreen}{HTML}{577D51}

\newcommand{\casefigure}{%
\begin{figure}[!htbp]\centering
\includegraphics[width=\linewidth]{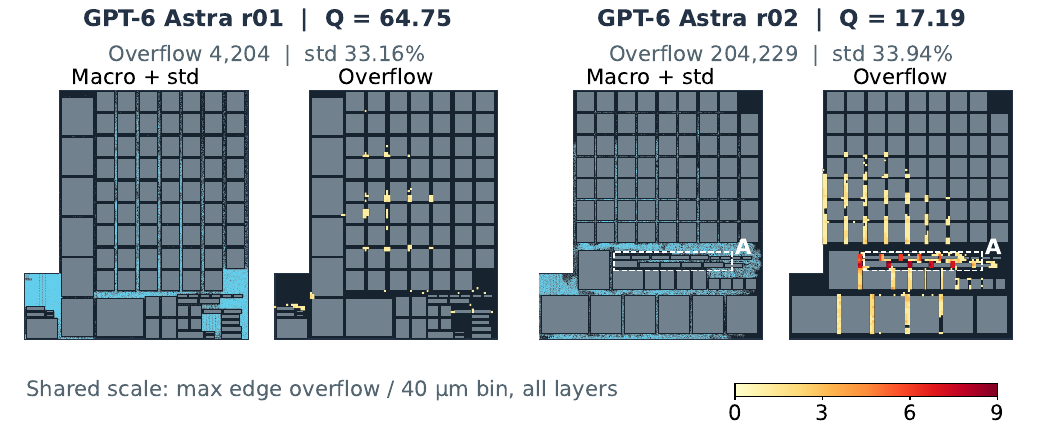}
\caption{\bi{同一模型与harness的两份已测布局。每对左图合并macro与实际std，右图显示所有层每40\,$\mu$m格的最大正overflow，共用0--9线性色阶，不可与全景图按颜色直接比值。r02的A区有横向小宏分隔的std条带，若干入口与热点重合；r01无同等强度热点。两例99颗macro均通过几何检查。图由原生placement后状态导出重绘；它定位差异，不单独证明通道宽度的因果作用。}{Two measured layouts under the same model and harness. Each pair combines macros and actual std at left, with maximum positive edge overflow across layers per 40\,$\mu$m bin at right on a shared linear 0--9 scale; colors cannot be compared numerically with the full-layout panels. In r02, region A contains std strips separated by horizontal small macros; several entrances coincide with hotspots absent at the same intensity in r01. Both sets of 99 macros pass geometry checks. Reconstructed from native post-placement exports: this locates differences without isolating a causal effect of channel width.}}
\label{fig:cases}\end{figure}}

\newcommand{\scoreequations}{%
\begin{align}
d(z;h)&=(1+z/h)^{-1},\label{eq:decay}\\
c&=d(p;0.0004)^{0.4}d(O;2000)^{0.4}d(M;20)^{0.2},\\
t_j&=d(W_j;2\,\mathrm{ns})^{0.4}d(N_j;200\,\mathrm{ns})^{0.4}d(V_j;2000)^{0.2},\\
t&=t_{rr}^{8/17}t_{rm}^{4/17}t_{mr}^{4/17}t_{other}^{1/17},\\
l&=d(\ell;180\,\mathrm{m}),\qquad u=\frac{A_{std}}{A_{die}-A_{macro}},\qquad s=1-u,\\
Q&=100\,c^{0.35}t^{0.50}l^{0.10}s^{0.05}.\label{eq:quality}
\end{align}}

\newcommand{\novafigure}{%
\begin{figure}[!p]\centering
\includegraphics[width=\linewidth]{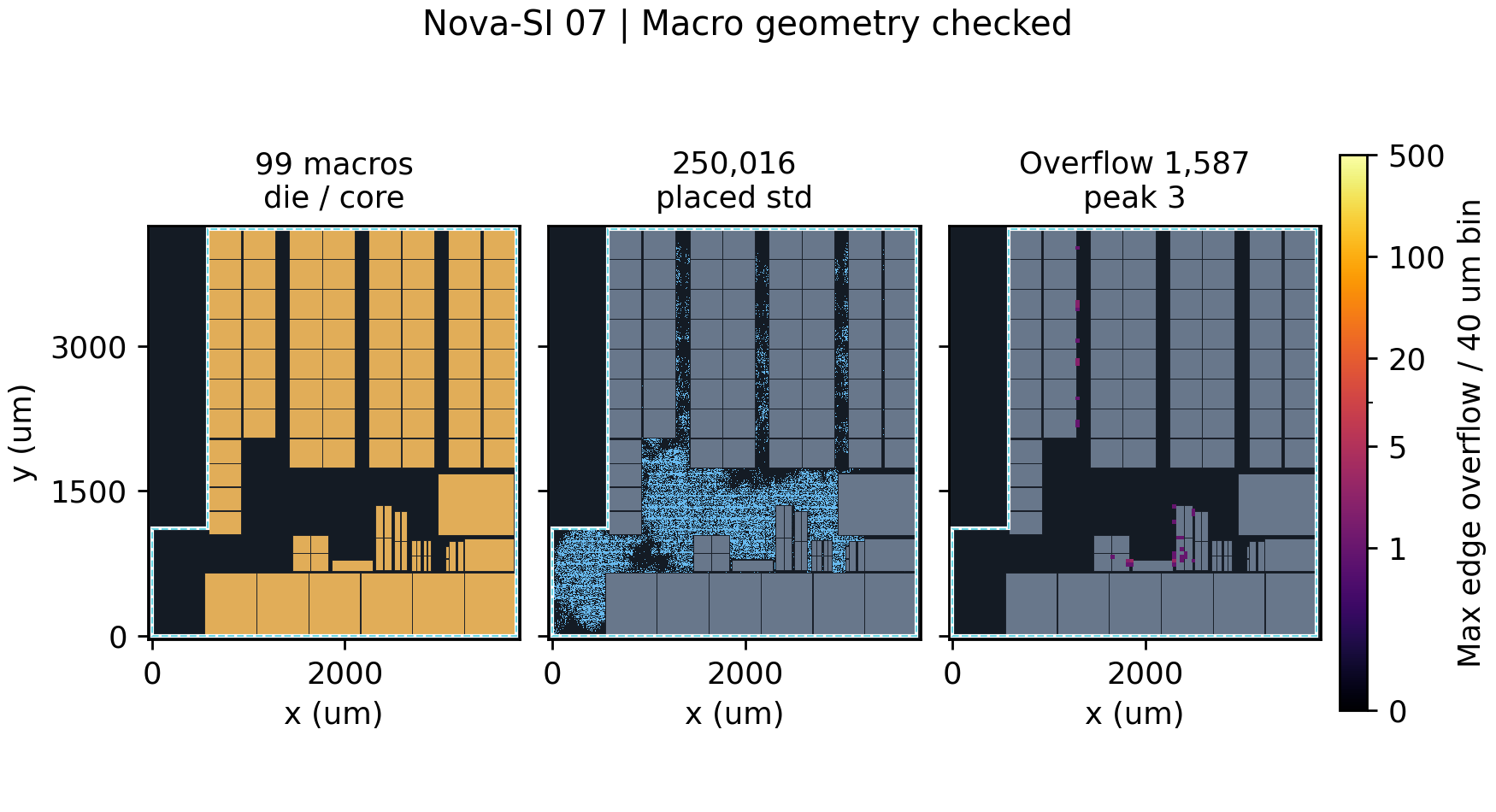}\\[-3pt]
\includegraphics[width=\linewidth]{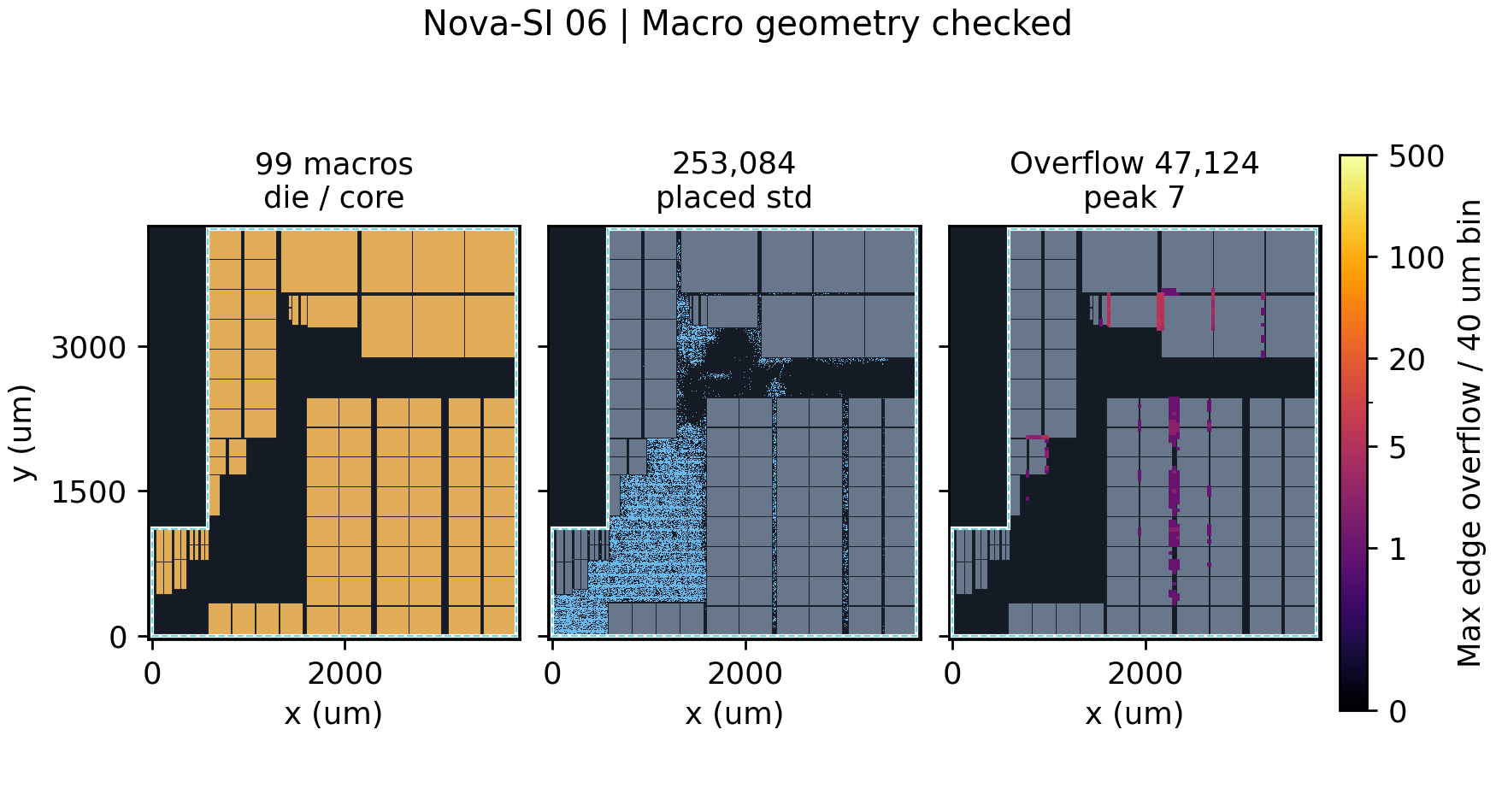}
\caption{\bi{Nova-SI最高分（07，$Q=81.09$）与最低分（06，$Q=39.01$）布局。两者是不同产物，不构成优化时间序列。每行依次为macro、实际std和同次测量的拥塞；每40\,$\mu$m格取跨层最大正overflow，共用0--500色阶，0--1线性、之后对数。07有连接下部std区域的竖向空间；06也有宽横向空白，却在宏阵列通道内出现更广热点。可见，留白总量不足以描述局部拥塞。白线为die，青色虚线为core；图由原生坐标和测量导出绘制。时序不由热图推断，另见表~\ref{tab:timing-case}。}{Nova-SI layouts with the highest (07, $Q=81.09$) and lowest (06, $Q=39.01$) measured scores. These are separate outputs, not successive optimization steps. Each row shows macros, actual std cells, and congestion from the same measurement: maximum positive overflow across layers per 40\,$\mu$m bin, on a shared 0--500 scale, linear over 0--1 and logarithmic thereafter. In 07, vertical open regions connect to the lower std region. Output 06 also leaves broad horizontal whitespace, yet has more extensive hotspots within macro-array channels. Whitespace alone does not characterize the local congestion. White denotes die and dashed cyan core; panels are reconstructed from native coordinate and measurement exports. Timing is measured separately in Table~\ref{tab:timing-case}, not inferred from the heatmap.}}
\label{fig:nova-cases}\end{figure}}

\newcommand{\invalidfigure}[1][!htbp]{%
\begin{figure}[#1]\centering
\includegraphics[width=\linewidth]{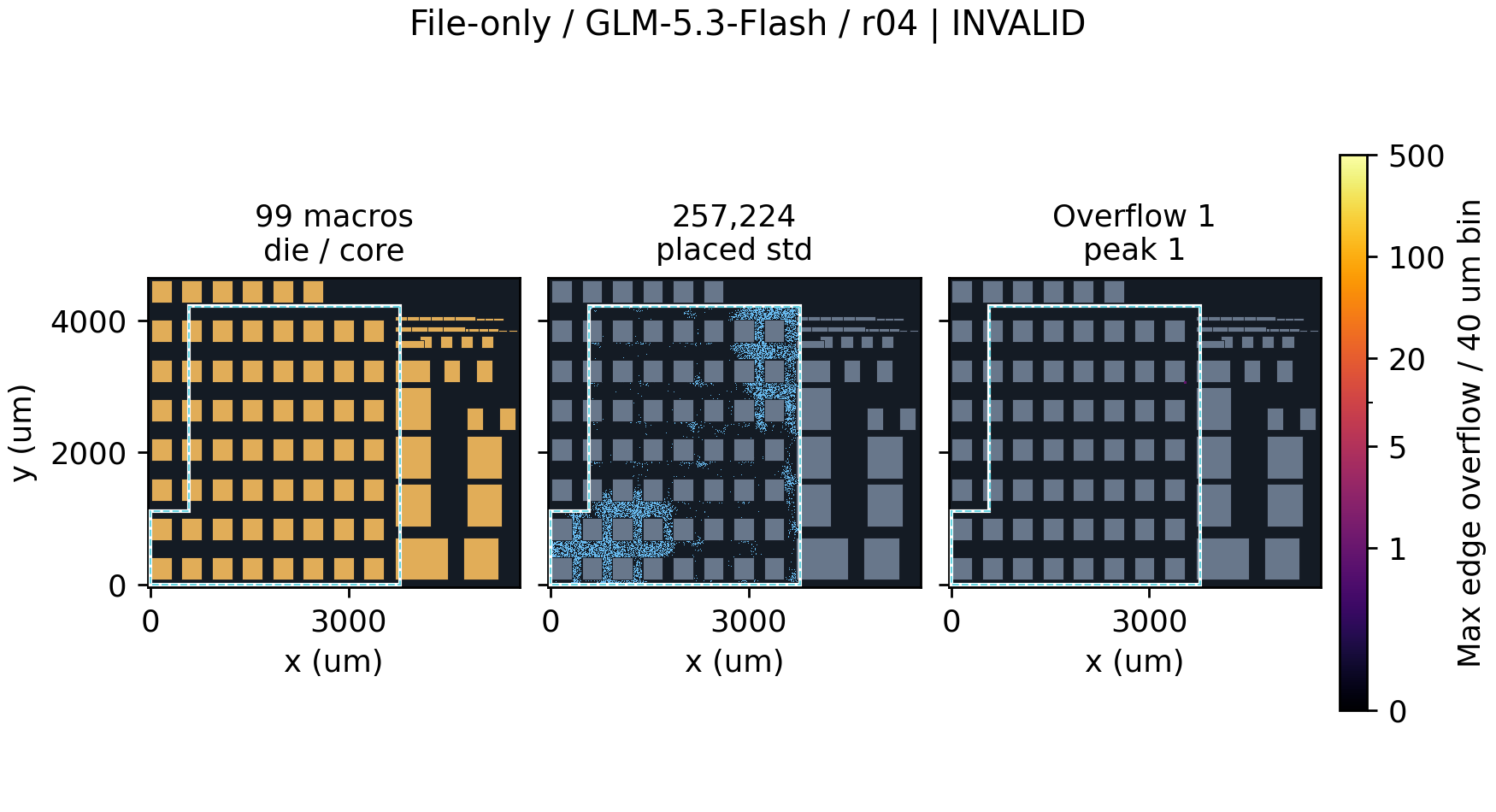}
\caption{\bi{低overflow并不总是好布局：File-only/GLM-5.3-Flash r04。坐标范围覆盖全部macro，包括白色die边界之外的对象。原生报告的overflow为1，但几何审查记录53个功能宏越出die，故有效分为0。该例的低overflow与严重几何违例并存，尚不足以确定低overflow的具体成因。三联图采用0--500色阶。}{Low overflow need not indicate a good layout: File-only/GLM-5.3-Flash r04. The axes cover all macros, including those outside the white die boundary. Native overflow is 1, but the geometry audit identifies 53 functional macros outside the die, giving effective score zero. Low overflow coexists with severe geometric violations here; the figure alone does not identify the specific cause of low overflow. The triptych uses a 0--500 color scale.}}
\label{fig:invalid}\end{figure}}

\newcommand{\qualitytable}{%
\begin{table}[!htbp]\centering\papertablestyle\setlength{\tabcolsep}{1.2pt}
\caption{\bi{质量主表与资源概览。质量项越高越好；几何非法与交付失败计0。}{Quality comparison with resource context. Higher quality is better; invalid geometry and delivery failure score zero.}}
\label{tab:quality}
\begin{tabular*}{\linewidth}{@{\extracolsep{\fill}}lccccccccc@{}}\toprule
&&\multicolumn{2}{c}{\tgroup{\bi{成绩与几何}{Outcomes}}}&\multicolumn{2}{c}{\tgroup{\bi{资源}{Resources}}}&\multicolumn{4}{c}{\tgroup{\bi{质量}{Quality} $\uparrow$}}\\
\cmidrule(lr){3-4}\cmidrule(lr){5-6}\cmidrule(lr){7-10}
\thd{\bi{系统／模型}{System / model}}&\thd{Harness}&\thd{$n/N$}&\thd{Inv.}&\thd{\thunit{Tok.}{(M)}}&\thd{\bi{核时}{Core-h}}&\thd{$C$}&\thd{$T$}&\thd{$Q_{\mathrm{max}}$}&\thd{$Q_{\mathrm{mean}}$}\\\midrule
\bi{人工参考$^\dagger$}{Human reference$^\dagger$}&---&1/1&---&---&---&\tnum{\textbf{94.17}}{94.17}&\tnum{\textbf{93.51}}{93.51}&\tnum{\textbf{91.88}}{91.88}&\tnum{\textbf{91.88}}{91.88}\\\midrule
Nova-SI&---&8/8&0&\tnum{131.36}{36.13}&\tnum{22.49}{22.49}&\tnum{\textbf{94.17}}{\textbf{34.67}}&\tnum{\textbf{93.51}}{\textbf{79.68}}&\tnum{\textbf{91.88}}{\textbf{81.09}}&\tnum{\textbf{91.88}}{\textbf{57.97}}\\
\addlinespace[3pt]\multicolumn{10}{l}{\textit{General}}\\
GPT-6 Astra\,{\scriptsize[xhigh]}&Codex&4/4&0&\tnum{131.36}{131.36}&\tnum{22.49}{30.79}&\tnum{\textbf{94.17}}{\underline{13.87}}&\tnum{\textbf{93.51}}{\underline{59.45}}&\tnum{\textbf{91.88}}{\underline{64.75}}&\tnum{\textbf{91.88}}{\underline{31.17}}\\
GPT-6.1 Sol\,{\scriptsize[xhigh]}&Codex&4/4&0&\tnum{131.36}{286.50}&\tnum{22.49}{53.81}&\tnum{\textbf{94.17}}{7.59}&\tnum{\textbf{93.51}}{39.89}&\tnum{\textbf{91.88}}{38.37}&\tnum{\textbf{91.88}}{22.73}\\
DeepSeek-V4.1-Flash-1017\,{\scriptsize[max]}&DeepSeek Harness&4/4&0&\tnum{131.36}{156.45}&\tnum{22.49}{25.89}&\tnum{\textbf{94.17}}{4.44}&\tnum{\textbf{93.51}}{34.58}&\tnum{\textbf{91.88}}{39.00}&\tnum{\textbf{91.88}}{16.18}\\
GLM-5.3\,{\scriptsize[max]}&zcode&4/4&0&\tnum{131.36}{62.11}&\tnum{22.49}{22.23}&\tnum{\textbf{94.17}}{3.23}&\tnum{\textbf{93.51}}{44.49}&\tnum{\textbf{91.88}}{22.71}&\tnum{\textbf{91.88}}{18.92}\\
GLM-5.3-Flash\,{\scriptsize[max]}&zcode&4/4&0&\tnum{131.36}{185.44}&\tnum{22.49}{76.80}&\tnum{\textbf{94.17}}{2.92}&\tnum{\textbf{93.51}}{36.57}&\tnum{\textbf{91.88}}{27.88}&\tnum{\textbf{91.88}}{15.52}\\
\addlinespace[3pt]\multicolumn{10}{l}{\textit{File-only}}\\
DeepSeek-V4.1-Flash-1017\,{\scriptsize[max]}&InternBootcamp&4/4&4&\tnum{131.36}{13.53}&\tnum{22.49}{20.10}&\tnum{\textbf{94.17}}{0.00}&\tnum{\textbf{93.51}}{0.00}&\tnum{\textbf{91.88}}{0.00}&\tnum{\textbf{91.88}}{0.00}\\
GLM-5.3\,{\scriptsize[max]}&InternBootcamp&4/4&1&\tnum{131.36}{9.64}&\tnum{22.49}{17.52}&\tnum{\textbf{94.17}}{5.27}&\tnum{\textbf{93.51}}{29.94}&\tnum{\textbf{91.88}}{43.39}&\tnum{\textbf{91.88}}{15.25}\\
GLM-5.3-Flash\,{\scriptsize[max]}&InternBootcamp&4/4&4&\tnum{131.36}{10.96}&\tnum{22.49}{15.32}&\tnum{\textbf{94.17}}{0.00}&\tnum{\textbf{93.51}}{0.00}&\tnum{\textbf{91.88}}{0.00}&\tnum{\textbf{91.88}}{0.00}\\
\bottomrule\end{tabular*}
\par\smallskip\begin{minipage}{\linewidth}\footnotesize\raggedright
\bi{$C$为拥塞分，$T$为时序分，质量均为百分制。$n/N$为已确定成绩数／纳入数，Inv.为宏几何非法的产物数。39份产物有完整测量；General/DeepSeek r03未完成placement，作为交付失败计0。除$Q_{\mathrm{max}}$外，质量项为全部$N$份结果的均值；几何非法与交付失败的板块分和总分均记0，原始物理量不补0。粗体／下划线标记Agent配置间最高／次高观测质量，REF不参与排名；不表示统计显著。}{\(C\): congestion; $T$: timing; all quality scores use a 0--100 scale. $n/N$: scored / included outcomes; Inv.: number of outputs with invalid macro geometry. Complete measurements exist for 39 outputs; General/DeepSeek r03 failed to complete placement and scores zero. Quality columns average all $N$ outcomes except $Q_{\mathrm{max}}$. Invalid geometry and delivery failure zero both components and total quality, not raw physical quantities. Bold / underline mark the highest / second-highest observed agent scores, excluding REF; not statistical significance.}\par
\bi{方括号为推理档位（附录~\ref{app:setting}），Tok.为总token。$Q_{\mathrm{max}}$仅取选定产物，不取trace历史最优。资源均值使用全部$N$份研究对象：Nova与File-only核时覆盖外评，General覆盖已记录的求解与评价；日志值为各自覆盖范围内的下界，不代表完整系统成本。$^\dagger$历史人工placement参考，非匹配重跑或signoff；不计入Agent样本，成本未知；横线不表示0。}{Brackets: reasoning effort (Appendix~\ref{app:setting}); Tok.: total tokens. $Q_{\mathrm{max}}$ uses selected outputs, not trace-wide best candidates. Resources average all $N$: Nova and File-only core-hours cover external evaluation; General covers recorded solving and evaluation. Logged values are lower bounds within these scopes, not full-system costs. $^\dagger$Historical engineer placement reference, not a matched rerun or signoff; excluded from agent cohorts; costs unknown; dashes do not denote zero.}
\end{minipage}\end{table}}
\newcommand{\resourcetable}{%
\begin{table}[!htbp]\centering\papertablestyle
\caption{\bi{对照配置的资源分项与PG+placement尝试次数均值。}{Mean resources and PG+placement attempts by baseline configuration.}}
\label{tab:resources}\begin{tabular*}{\linewidth}{@{\extracolsep{\fill}}lccccc@{}}\toprule
\thd{\bi{系统／模型}{System / model}}&\thd{\bi{输入token}{Input tokens} (M)}&\thd{\bi{输出token}{Output tokens} (M)}&\thd{EDA (core-h)}&\thd{\bi{尝试次数}{Attempts}}&\thd{\bi{次数有记录/$N$}{Counted/$N$}}\\\midrule
\bi{人工参考$^\dagger$}{Human reference$^\dagger$}&---&---&---&---&---\\\midrule
\addlinespace[3pt]\multicolumn{6}{l}{\textit{General}}\\
GPT-6 Astra&\tnum{130.729}{130.729}&\tnum{0.631}{0.631}&\tnum{30.79}{30.79}&\tnum{2.25}{2.25}&4/4\\
GPT-6.1 Sol&\tnum{130.729}{285.621}&\tnum{0.631}{0.883}&\tnum{30.79}{53.81}&\tnum{2.25}{2.75}&4/4\\
DeepSeek-V4.1-Flash-1017&\tnum{130.729}{156.154}&\tnum{0.631}{0.292}&\tnum{30.79}{25.89}&\tnum{2.25}{2.00}&3/4\\
GLM-5.3&\tnum{130.729}{61.486}&\tnum{0.631}{0.622}&\tnum{30.79}{22.23}&\tnum{2.25}{4.25}&4/4\\
GLM-5.3-Flash&\tnum{130.729}{185.211}&\tnum{0.631}{0.234}&\tnum{30.79}{76.80}&\tnum{2.25}{6.00}&4/4\\
\addlinespace[3pt]\multicolumn{6}{l}{\textit{File-only}}\\
DeepSeek-V4.1-Flash-1017&\tnum{130.729}{13.413}&\tnum{0.631}{0.114}&\tnum{30.79}{20.10}&\tnum{2.25}{1.50}&4/4\\
GLM-5.3&\tnum{130.729}{9.462}&\tnum{0.631}{0.182}&\tnum{30.79}{17.52}&\tnum{2.25}{1.50}&4/4\\
GLM-5.3-Flash&\tnum{130.729}{10.854}&\tnum{0.631}{0.102}&\tnum{30.79}{15.32}&\tnum{2.25}{1.25}&4/4\\
\bottomrule\end{tabular*}
\par\smallskip\begin{minipage}{\linewidth}\footnotesize\raggedright
\bi{每个对照配置$N=4$。尝试次数为计数已知运行的PG+placement调用均值（含失败）；次数有记录/$N$为有尝试计数的运行数／纳入数，缺失不作0。General在线调用，File-only外评且无在线调用。核时为日志CPU秒数/3600（下界），非申请核数乘墙钟。$^\dagger$人工成本与次数未知。}{Each baseline has $N=4$. Attempts averages recorded PG+placement calls, including failures, over counted runs; Counted/$N$ gives runs with attempt counts / included runs. Missing counts are not zero. Counts are online for General and external for File-only (no online calls). Core-hours are logged CPU seconds / 3600, a lower bound, not reserved cores times wall time. $^\dagger$Human costs and attempts are unknown.}
\end{minipage}\end{table}}
\newcommand{\novacosttable}{%
\begin{table}[!htbp]\centering\papertablestyle
\caption{\bi{Nova-SI逐产物资源记录。EDA核时覆盖外部评价。}{Per-output Nova-SI resource records. EDA hours come from external evaluation.}}
\label{tab:nova-cost}\begin{tabular*}{\linewidth}{@{\extracolsep{\fill}}lcccc@{}}\toprule
\thd{ID}&\thd{\bi{输入token}{Input tokens} (M)}&\thd{\bi{输出token}{Output tokens} (M)}&\thd{\bi{总token}{Total tokens} (M)}&\thd{EDA (core-h)}\\\midrule
01&\tnum{20.161}{20.161}&\tnum{11.247}{8.349}&\tnum{28.510}{28.510}&\tnum{17.00}{17.00}\\
02&\tnum{20.161}{22.184}&\tnum{11.247}{11.247}&\tnum{28.510}{33.431}&\tnum{17.00}{19.37}\\
03&\tnum{20.161}{19.402}&\tnum{11.247}{19.866}&\tnum{28.510}{39.268}&\tnum{17.00}{19.29}\\
04&\tnum{20.161}{18.769}&\tnum{11.247}{12.818}&\tnum{28.510}{31.587}&\tnum{17.00}{26.71}\\
05&\tnum{20.161}{9.099}&\tnum{11.247}{9.000}&\tnum{28.510}{18.099}&\tnum{17.00}{23.89}\\
06&\tnum{20.161}{21.287}&\tnum{11.247}{13.790}&\tnum{28.510}{35.077}&\tnum{17.00}{21.91}\\
07&\tnum{20.161}{18.911}&\tnum{11.247}{18.596}&\tnum{28.510}{37.507}&\tnum{17.00}{22.38}\\
08&\tnum{20.161}{32.603}&\tnum{11.247}{32.940}&\tnum{28.510}{65.543}&\tnum{17.00}{29.34}\\
\bottomrule\end{tabular*}
\par\smallskip\begin{minipage}{\linewidth}\footnotesize\raggedright
\bi{\raggedright Token按被测产物对应记录统计；原生响应计数包含日志中有记录的被拒绝响应，reasoning token不在输出之外重复累加。上游记录可能重叠，逐产物总和不等于去重的独立请求总量。EDA仅覆盖外部评价。}{Tokens use records associated with each measured output. Native-response accounting includes rejected responses recorded in the available logs and does not add reasoning tokens again outside completion tokens. Upstream records may overlap; summing outputs does not establish a deduplicated request total. EDA covers external evaluation only.}
\end{minipage}\end{table}}
\newcommand{\novatable}{%
\begin{table}[!htbp]\centering\papertablestyle
\caption{\bi{Nova-SI逐产物质量与历史人工参考。}{Per-output Nova-SI quality and the historical human reference.}}
\label{tab:nova-cases}\begin{tabular*}{\linewidth}{@{\extracolsep{\fill}}lcccccc@{}}\toprule
\thd{ID}&\thd{Overflow}&\thd{Peak}&\thd{std (\%)}&\thd{$C\uparrow$}&\thd{$T\uparrow$}&\thd{$Q\uparrow$}\\\midrule
REF&\tnum{10,465}{142}&2&\tnum{32.13}{32.13}&\tnum{\textbf{94.17}}{94.17}&\tnum{\textbf{93.51}}{93.51}&\tnum{\textbf{91.88}}{91.88}\\\midrule
01&\tnum{10,465}{9,186}&9&\tnum{32.13}{32.22}&\tnum{\textbf{94.17}}{31.42}&\tnum{\textbf{93.51}}{81.66}&\tnum{\textbf{91.88}}{58.42}\\
02&\tnum{10,465}{10,465}&12&\tnum{32.13}{32.19}&\tnum{\textbf{94.17}}{28.18}&\tnum{\textbf{93.51}}{82.76}&\tnum{\textbf{91.88}}{56.62}\\
03&\tnum{10,465}{7,381}&10&\tnum{32.13}{32.30}&\tnum{\textbf{94.17}}{35.37}&\tnum{\textbf{93.51}}{80.33}&\tnum{\textbf{91.88}}{60.39}\\
04&\tnum{10,465}{34,981}&20&\tnum{32.13}{32.30}&\tnum{\textbf{94.17}}{16.09}&\tnum{\textbf{93.51}}{65.28}&\tnum{\textbf{91.88}}{41.33}\\
05&\tnum{10,465}{27,867}&12&\tnum{32.13}{32.26}&\tnum{\textbf{94.17}}{20.12}&\tnum{\textbf{93.51}}{\underline{88.62}}&\tnum{\textbf{91.88}}{52.07}\\
06&\tnum{10,465}{47,124}&7&\tnum{32.13}{32.48}&\tnum{\textbf{94.17}}{13.24}&\tnum{\textbf{93.51}}{66.73}&\tnum{\textbf{91.88}}{39.01}\\
07&\tnum{10,465}{1,587}&3&\tnum{32.13}{32.16}&\tnum{\textbf{94.17}}{\textbf{68.96}}&\tnum{\textbf{93.51}}{\textbf{90.69}}&\tnum{\textbf{91.88}}{\textbf{81.09}}\\
08&\tnum{10,465}{1,913}&3&\tnum{32.13}{32.21}&\tnum{\textbf{94.17}}{\underline{63.99}}&\tnum{\textbf{93.51}}{81.41}&\tnum{\textbf{91.88}}{\underline{74.84}}\\
\bottomrule\end{tabular*}
\tablenote{\bi{01--08按产物编号排列，均有99颗macro并通过边界与互不重叠检查。Overflow和Peak为计数，std为扣除宏面积后计算的标准单元利用率；$C,T,Q$分别为拥塞、setup时序与总质量的百分制分数，粗体／下划线标记八份产物中最高／次高分，同显示值并列。REF不参与排名或Nova统计；它是历史placement参考，非匹配重跑或signoff。}{Outputs 01--08 follow output-ID order; all 99 macros pass boundary and overlap checks. Overflow and Peak are counts; std denotes standard-cell utilization, with macro area excluded from its denominator. Congestion $C$, setup timing $T$, and total quality $Q$ are scores from 0 to 100; bold / underline mark the highest / second-highest among the eight outputs, with printed ties sharing a mark. REF is excluded from ranking and Nova statistics: it is a historical placement reference, not a matched rerun or signoff.}}
\end{table}}

\newcommand{\physicaltable}{%
\begin{table}[!htbp]\centering\papertablestyle
\caption{\bi{宏几何通过布局的原始物理量均值。}{Mean raw physical quantities for geometry-passing layouts.}}
\label{tab:physical}\begin{tabular*}{\linewidth}{@{\extracolsep{\fill}}lcccccc@{}}\toprule
\thd{\bi{系统／模型}{System / model}}&\thd{$n_g$}&\thd{Overflow}&\thd{Peak}&\thd{GRC (\%)}&\thd{\bi{线长}{Length} (m)}&\thd{std (\%)}\\\midrule
\bi{人工参考$^\dagger$}{Human reference$^\dagger$}&1&\tnum{1,033,158}{142}&\tnum{23.25}{2.00}&\tnum{0.0014}{0.0014}&\tnum{20.31}{20.31}&\tnum{32.13}{32.13}\\\midrule
Nova-SI&8&\tnum{1,033,158}{17,563}&\tnum{23.25}{9.50}&\tnum{0.0014}{0.0745}&\tnum{20.31}{21.89}&\tnum{32.13}{32.27}\\
\addlinespace[3pt]\multicolumn{7}{l}{\textit{General}}\\
GPT-6 Astra&4&\tnum{1,033,158}{281,748}&\tnum{23.25}{23.25}&\tnum{0.0014}{0.9586}&\tnum{20.31}{28.88}&\tnum{32.13}{33.56}\\
GPT-6.1 Sol&4&\tnum{1,033,158}{526,552}&\tnum{23.25}{19.50}&\tnum{0.0014}{1.7251}&\tnum{20.31}{31.11}&\tnum{32.13}{33.80}\\
DeepSeek-V4.1-Flash-1017&3&\tnum{1,033,158}{673,784}&\tnum{23.25}{43.33}&\tnum{0.0014}{1.5112}&\tnum{20.31}{27.29}&\tnum{32.13}{33.39}\\
GLM-5.3&4&\tnum{1,033,158}{291,349}&\tnum{23.25}{29.50}&\tnum{0.0014}{1.3516}&\tnum{20.31}{30.31}&\tnum{32.13}{33.82}\\
GLM-5.3-Flash&4&\tnum{1,033,158}{1,033,158}&\tnum{23.25}{49.75}&\tnum{0.0014}{4.0209}&\tnum{20.31}{35.46}&\tnum{32.13}{35.15}\\
\addlinespace[3pt]\multicolumn{7}{l}{\textit{File-only}}\\
GLM-5.3&3&\tnum{1,033,158}{1,365,605}&\tnum{23.25}{78.00}&\tnum{0.0014}{3.7338}&\tnum{20.31}{31.96}&\tnum{32.13}{35.05}\\
\bottomrule\end{tabular*}
\tablenote{\bi{$n_g$为几何通过的已测布局数；没有此类布局的配置不列出。非法产物的原始物理量不填0。Overflow与Peak为计数，GRC指全局布线网格；GRC (\%)为发生overflow的网格百分比，std为扣除宏面积后计算的标准单元利用率。}{\(n_g\) counts measured geometry-passing layouts; configurations with none are omitted. Invalid physical quantities are not imputed as zero. Overflow and Peak are counts. GRC denotes a global-routing cell; GRC (\%) reports the percentage of these cells with overflow. The std column denotes standard-cell utilization, with macro area excluded from its denominator.}}
\par\smallskip\begin{minipage}{\linewidth}\footnotesize\raggedright
\bi{$^\dagger$人工参考是单份历史placement观测，不是匹配重跑或signoff；不计入Agent样本统计。}{\(\dagger\) Single historical engineer placement reference, not a matched rerun or signoff; excluded from agent-cohort statistics.}
\end{minipage}\end{table}}
\newcommand{\pairedtable}{%
\begin{table}[!htbp]\centering\papertablestyle
\caption{\bi{首次完整测量与选定布局的配对质量。}{Quality at the first complete measurement and the selected layout.}}
\label{tab:paired}\begin{tabular*}{\linewidth}{@{\extracolsep{\fill}}lcccc@{}}\toprule
\thd{\bi{系统／模型}{System / model}}&\thd{$n_g$}&\thd{$Q_{\mathrm{first}}$}&\thd{$Q_{\mathrm{selected}}$}&\thd{$\Delta Q$}\\\midrule
Nova-SI&8&\tnum{57.97}{57.97}&\tnum{57.97}{57.97}&\tnum{$+$0.00}{$+$0.00}\\
\addlinespace[3pt]\multicolumn{5}{l}{\textit{General}}\\
GPT-6 Astra&4&\tnum{57.97}{31.17}&\tnum{57.97}{31.17}&\tnum{$+$0.00}{$+$0.00}\\
GPT-6.1 Sol&4&\tnum{57.97}{26.19}&\tnum{57.97}{22.73}&\tnum{$+$0.00}{$-$3.46}\\
DeepSeek-V4.1-Flash-1017&3&\tnum{57.97}{21.58}&\tnum{57.97}{21.58}&\tnum{$+$0.00}{$+$0.00}\\
GLM-5.3&4&\tnum{57.97}{18.80}&\tnum{57.97}{18.92}&\tnum{$+$0.00}{$+$0.12}\\
GLM-5.3-Flash&4&\tnum{57.97}{14.37}&\tnum{57.97}{15.52}&\tnum{$+$0.00}{$+$1.14}\\
\addlinespace[3pt]\multicolumn{5}{l}{\textit{File-only}}\\
GLM-5.3&3&\tnum{57.97}{20.33}&\tnum{57.97}{20.33}&\tnum{$+$0.00}{$+$0.00}\\
\bottomrule\end{tabular*}
\tablenote{\bi{$n_g$为几何通过的配对数，$Q$为百分制均值；$\Delta Q$由未舍入的选定值减首次值计算。首次完整测量不一定来自第一次尝试；选定测量对应交付布局而非历史最高分。Nova与File-only两端重复引用同一次提交后外评，不构成在线优化轨迹。}{\(n_g\) counts geometry-passing pairs; $Q$ is the mean on a 0--100 scale. $\Delta Q$ is selected minus first, computed before rounding. The first complete measurement need not come from the first attempt; selected refers to the delivered layout, not the historical best. Nova and File-only reuse the same external measurement at both endpoints, not an online optimization trajectory.}}
\end{table}}
\newcommand{\timingtable}{%
\begin{table}[!htbp]\centering\papertablestyle
\caption{\bi{宏几何通过布局的最坏场景setup时序均值。}{Mean worst-scenario setup timing for geometry-passing layouts.}}
\label{tab:timing}\begin{tabular*}{\linewidth}{@{\extracolsep{\fill}}lccccccc@{}}\toprule
&&\multicolumn{2}{c}{\tgroup{reg$\to$reg (ns)}}&\multicolumn{2}{c}{\tgroup{reg$\to$mem (ns)}}&\multicolumn{2}{c}{\tgroup{mem$\to$reg (ns)}}\\
\cmidrule(lr){3-4}\cmidrule(lr){5-6}\cmidrule(lr){7-8}
\thd{\bi{系统／模型}{System / model}}&\thd{$n_g$}&\thd{WNS}&\thd{TNS}&\thd{WNS}&\thd{TNS}&\thd{WNS}&\thd{TNS}\\\midrule
\bi{人工参考$^\dagger$}{Human reference$^\dagger$}&1&\tnum{$-$0.229}{$-$0.229}&\tnum{$-$3862.730}{$-$15.105}&\tnum{$-$0.209}{$-$0.209}&\tnum{$-$1267.639}{$-$0.735}&\tnum{$-$0.099}{$-$0.099}&\tnum{$-$109.591}{$-$0.526}\\\midrule
Nova-SI&8&\tnum{$-$0.229}{$-$1.225}&\tnum{$-$3862.730}{$-$44.773}&\tnum{$-$0.209}{$-$1.104}&\tnum{$-$1267.639}{$-$41.806}&\tnum{$-$0.099}{$-$0.803}&\tnum{$-$109.591}{$-$10.831}\\
\addlinespace[3pt]\multicolumn{8}{l}{\textit{General}}\\
GPT-6 Astra&4&\tnum{$-$0.229}{$-$2.352}&\tnum{$-$3862.730}{$-$530.426}&\tnum{$-$0.209}{$-$2.240}&\tnum{$-$1267.639}{$-$157.122}&\tnum{$-$0.099}{$-$2.325}&\tnum{$-$109.591}{$-$85.630}\\
GPT-6.1 Sol&4&\tnum{$-$0.229}{$-$8.830}&\tnum{$-$3862.730}{$-$3862.730}&\tnum{$-$0.209}{$-$2.608}&\tnum{$-$1267.639}{$-$280.384}&\tnum{$-$0.099}{$-$3.326}&\tnum{$-$109.591}{$-$109.591}\\
DeepSeek-V4.1-Flash-1017&3&\tnum{$-$0.229}{$-$4.327}&\tnum{$-$3862.730}{$-$2841.980}&\tnum{$-$0.209}{$-$3.476}&\tnum{$-$1267.639}{$-$528.565}&\tnum{$-$0.099}{$-$2.156}&\tnum{$-$109.591}{$-$92.864}\\
GLM-5.3&4&\tnum{$-$0.229}{$-$3.367}&\tnum{$-$3862.730}{$-$797.730}&\tnum{$-$0.209}{$-$3.450}&\tnum{$-$1267.639}{$-$384.916}&\tnum{$-$0.099}{$-$3.483}&\tnum{$-$109.591}{$-$184.465}\\
GLM-5.3-Flash&4&\tnum{$-$0.229}{$-$3.276}&\tnum{$-$3862.730}{$-$1926.729}&\tnum{$-$0.209}{$-$6.441}&\tnum{$-$1267.639}{$-$1267.639}&\tnum{$-$0.099}{$-$3.709}&\tnum{$-$109.591}{$-$195.556}\\
\addlinespace[3pt]\multicolumn{8}{l}{\textit{File-only}}\\
GLM-5.3&3&\tnum{$-$0.229}{$-$6.757}&\tnum{$-$3862.730}{$-$2393.181}&\tnum{$-$0.209}{$-$3.224}&\tnum{$-$1267.639}{$-$603.979}&\tnum{$-$0.099}{$-$2.720}&\tnum{$-$109.591}{$-$132.878}\\
\bottomrule\end{tabular*}
\tablenote{\bi{$n_g$为几何通过的布局数。每个布局先在四场景分别取最坏WNS（最坏负slack）与TNS（总负slack），再跨布局平均；两者可来自不同场景。reg为寄存器，mem为存储器；数值越接近0，违例越轻。}{\(n_g\) counts geometry-passing layouts. WNS (worst negative slack) and TNS (total negative slack) are each minimized across four scenarios per layout, then averaged across layouts; the worst scenarios can differ. Here reg means register and mem means memory; values closer to zero indicate less severe violations.}}
\par\smallskip\begin{minipage}{\linewidth}\footnotesize\raggedright
\bi{$^\dagger$人工参考是单份历史placement观测，不是匹配重跑或signoff；不计入Agent样本统计。}{\(\dagger\) Single historical engineer placement reference, not a matched rerun or signoff; excluded from agent-cohort statistics.}
\end{minipage}\end{table}}
\newcommand{\coveragebounds}{%
\begin{table}[!htbp]\centering\papertablestyle
\caption{\bi{完整研究集合的物理测量与交付状态。}{Physical-measurement and delivery outcomes for the full cohort.}}
\label{tab:coverage-bounds}\begin{tabular*}{\linewidth}{@{\extracolsep{\fill}}lcccc@{}}\toprule
\thd{\bi{系统／模型}{System / model}}&\thd{$N$}&\thd{\bi{有测量}{Measured}}&\thd{Inv.}&\thd{\bi{交付失败}{Failed}}\\\midrule
Nova-SI&8&8&0&0\\
\addlinespace[3pt]\multicolumn{5}{l}{\textit{General}}\\
GPT-6 Astra&4&4&0&0\\
GPT-6.1 Sol&4&4&0&0\\
DeepSeek-V4.1-Flash-1017&4&3&0&1\\
GLM-5.3&4&4&0&0\\
GLM-5.3-Flash&4&4&0&0\\
\addlinespace[3pt]\multicolumn{5}{l}{\textit{File-only}}\\
DeepSeek-V4.1-Flash-1017&4&4&4&0\\
GLM-5.3&4&4&1&0\\
GLM-5.3-Flash&4&4&4&0\\
\bottomrule\end{tabular*}
\tablenote{\bi{Inv.为宏几何非法的已测产物，包含在测量列中；交付失败与其互斥。General/DeepSeek r03未完成placement，计0分，但物理测量保留为空。所有40份结果均有确定成绩，不需要对未决结果估计均值界。}{Inv. counts measured outputs with invalid macro geometry and is included in Measured; Failed is disjoint. General/DeepSeek r03 did not complete placement and scores zero, with physical measurements absent. All 40 outcomes have assigned scores; bounds for unresolved outcomes are unnecessary.}}
\end{table}}

\newcommand{\astraoverviewfigure}{%
\begin{figure}[!p]\centering
\includegraphics[width=\linewidth]{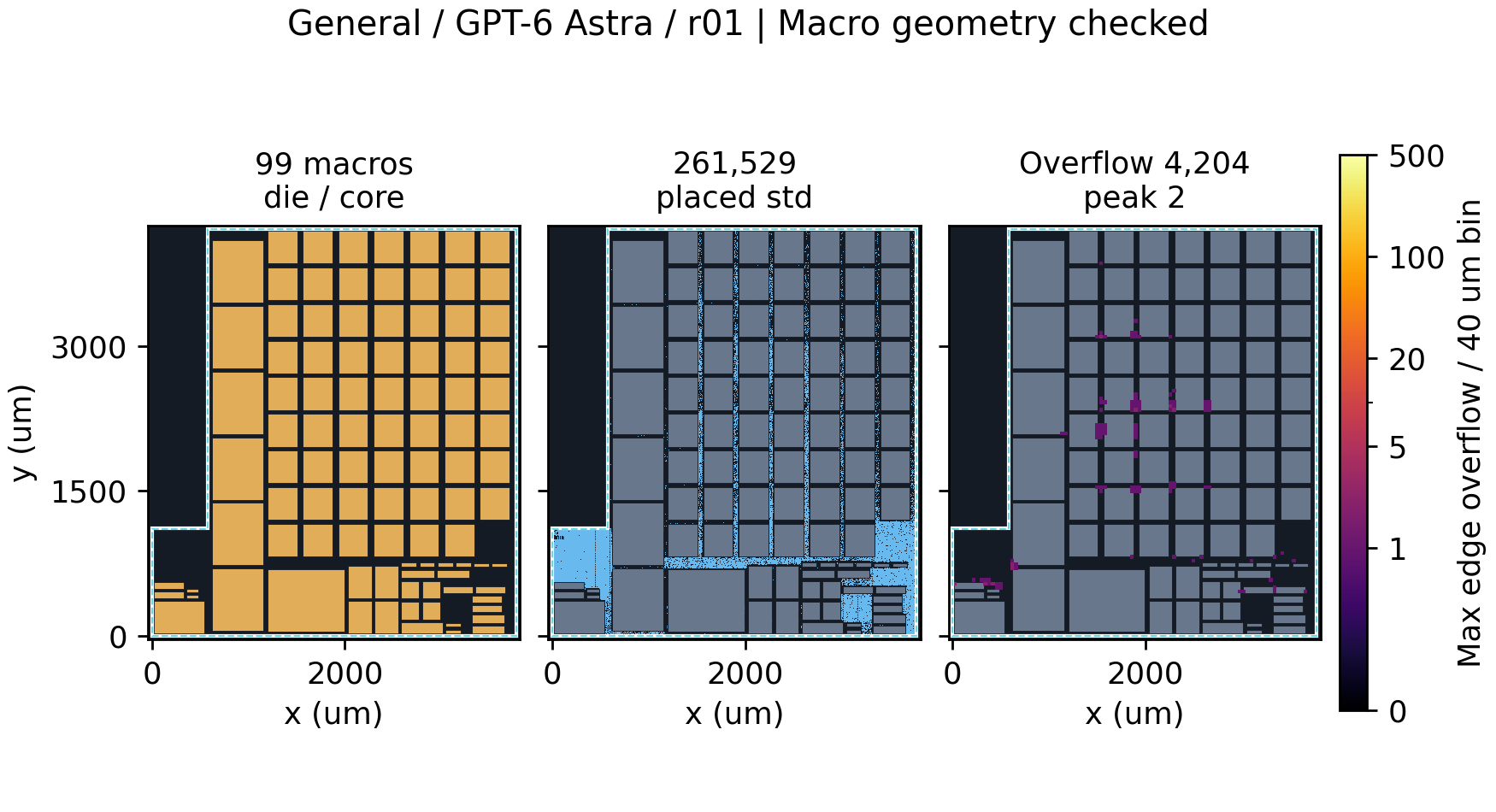}\\[-3pt]
\includegraphics[width=\linewidth]{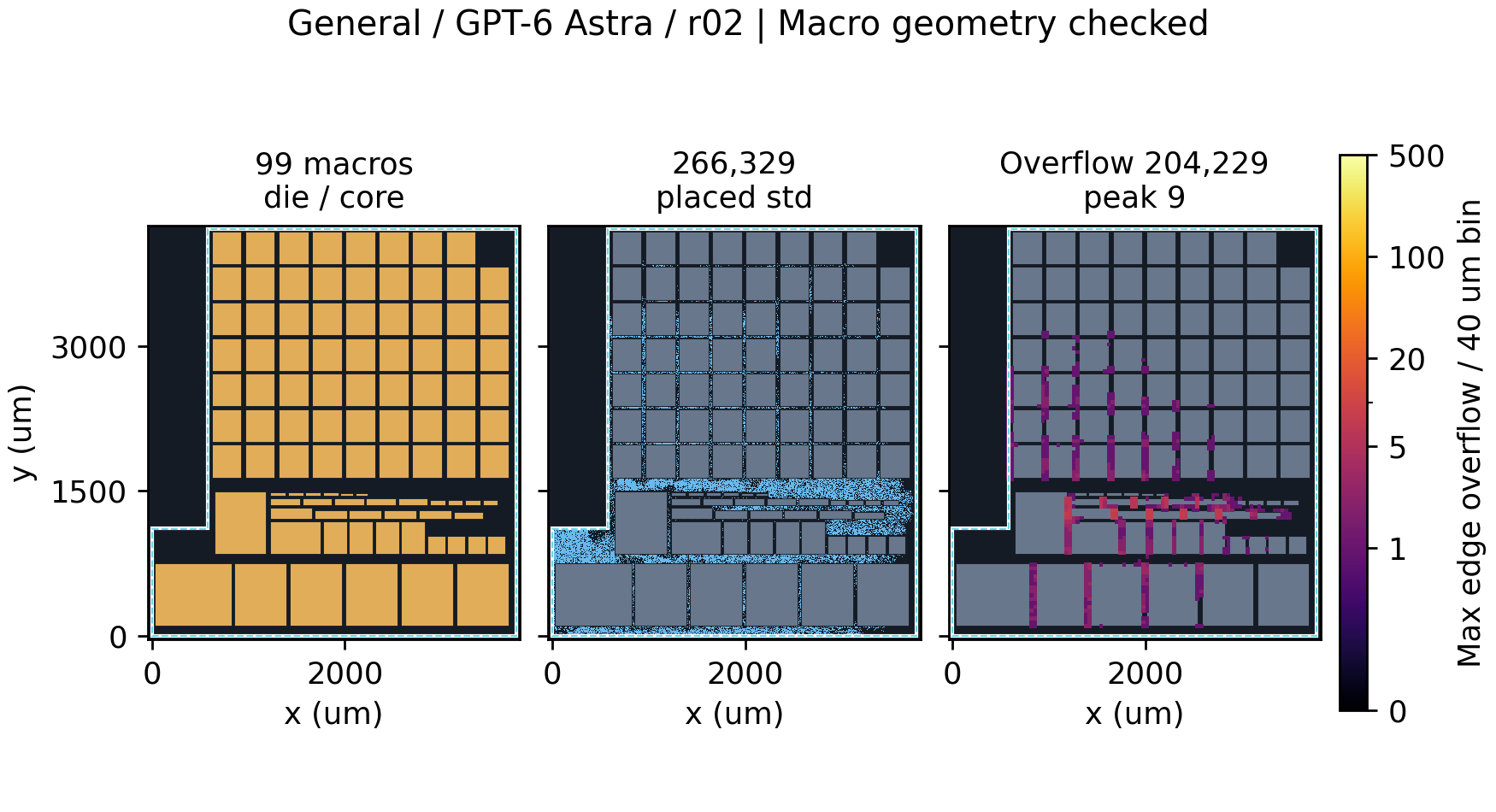}
\caption{\bi{General/GPT-6 Astra最高（r01，$Q=64.75$）与最低（r02，$Q=17.19$）选定产物，与图~\ref{fig:nova-cases}的Nova极值采用相同选例规则。左至右为macro、实际std和拥塞；共用0--500色阶，0--1线性、之后对数，每40\,$\mu$m格取跨层最大正overflow。两例全局std利用率为33.16\%与33.94\%，总overflow却为4,204与204,229。r02下部横向小宏之间的std条带和入口附近可见较广热点；r01拥塞较轻，其质量分高于六份Nova产物。白线为die，青色虚线为core；图由原生坐标和测量导出绘制。}{Highest (r01, $Q=64.75$) and lowest (r02, $Q=17.19$) selected General/GPT-6 Astra outputs, using the same extrema rule as Nova in Figure~\ref{fig:nova-cases}. Left to right: macros, actual std, and congestion. Both figures use a shared 0--500 scale, linear over 0--1 and logarithmic thereafter, with the maximum positive overflow across layers per 40\,$\mu$m bin. Global std utilization is 33.16\% versus 33.94\%, but total overflow is 4,204 versus 204,229. In r02, broader hotspots occur near std strips and entrances between lower horizontal macros. Output r01 is less congested and outperforms six Nova outputs. White denotes die and dashed cyan core; panels are reconstructed from native coordinate and measurement exports.}}
\label{fig:astra-overview}\end{figure}}

\newcommand{\taskfigure}{%
\begin{figure}[!htbp]\centering
\includegraphics[width=\linewidth]{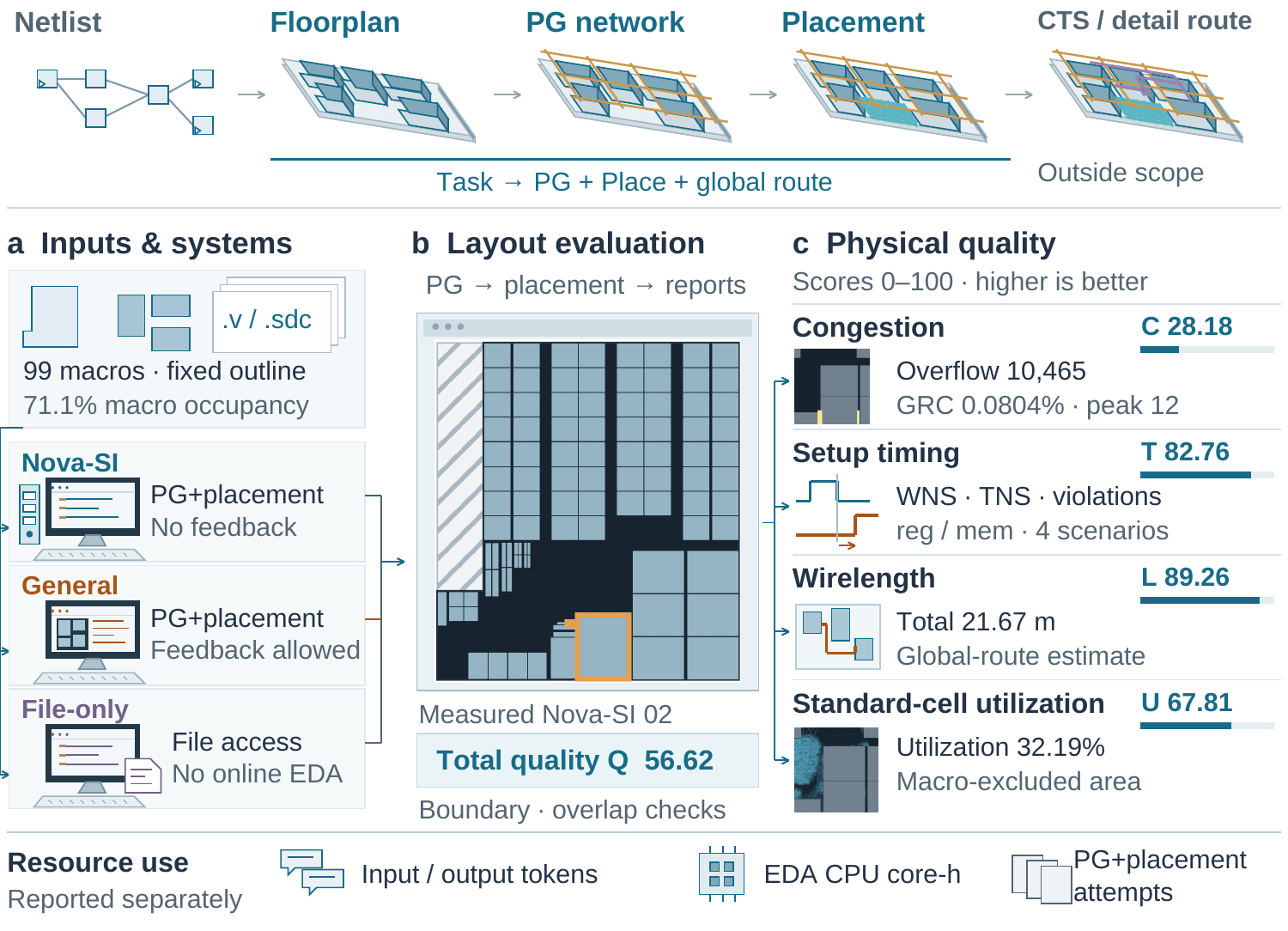}
\caption{\bi{从Agent交付到物理质量的评估框架。(a) 固定输入与三类系统接口，只有General允许在线PG+placement反馈。(b) 检查提交宏几何，并经电源与地（PG）网络构建、placement、全局布线估计和setup时序分析获得测量，不包含signoff。实际布局为Nova-SI产物02，而非共同输入；斜线区不可放置，橙框标出最大与最小宏。(c) 同一产物的拥塞、时序、线长和利用率得分（C/T/L/U，0--100分），资源另报。U为1减宏外面积中的std利用率再乘100，并非原始利用率百分比。拥塞/std局部图窗口相同；每40\,$\mu$m格取跨层最大正overflow，黄至红为0--12。顶部流程、工作站和时序波形均为示意，不是原生关键路径。评分见第~\ref{sec:metrics}节。}{Evaluation from agent deliverables to physical quality. (a) Fixed inputs and three interfaces; only General permits online PG+placement feedback. (b) Macro-geometry checks and power-and-ground (PG) network construction, placement, global-route estimation, and setup timing analysis connect submissions to measurements; signoff is excluded. The layout is Nova-SI output 02, not a common input. Hatching excludes placement; orange outlines mark the largest and smallest macros. (c) Congestion, timing, wirelength, and utilization scores (C/T/L/U, 0--100) for that output; resources are separate. U is 100 times one minus std utilization over macro-excluded area, not the raw utilization percentage. Congestion/std close-ups share a window; colors show maximum positive overflow across layers per 40\,$\mu$m bin, yellow--red at 0--12. Top stages, workstations, and timing waveforms are schematic, not native critical paths. Section~\ref{sec:metrics} defines scoring.}}
\label{fig:task}\end{figure}}

\newcommand{\conditionsummary}{%
\begin{center}\papertablestyle
\begin{tabular*}{\linewidth}{@{\extracolsep{\fill}}l p{125pt} p{165pt}@{}}
\toprule
\thd{\bi{系统条件}{Condition}} & \thd{\bi{求解接口}{Solving interface}} & \thd{\bi{规定PG+Place流程的在线反馈}{Online feedback from prescribed PG+Place}} \\
\midrule
Nova-SI & \bi{专用流水线}{Specialized pipeline} & \bi{不使用}{Not used} \\
General & \bi{原生工程工具}{Native engineering tools} & \bi{允许，不要求必须调用}{Allowed, not required} \\
File-only & \bi{文件操作与状态报告}{Files and status reporting} & \bi{不可用}{Unavailable} \\
\bottomrule
\end{tabular*}
\end{center}}

\newcommand{\distributionfigure}{%
\begin{figure}[!htbp]\centering
\includegraphics[width=\linewidth]{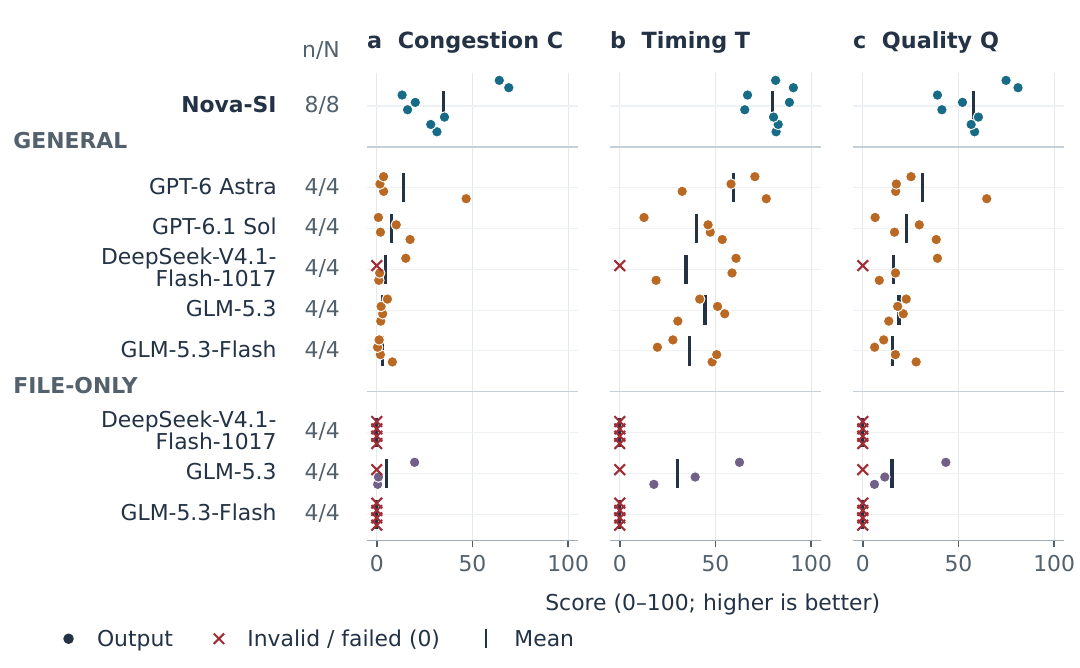}
\caption{\bi{逐产物拥塞、时序与质量分。三栏同尺度、同行配置与run错位一致。每栏含30份几何通过产物（彩色点）、9份几何非法产物及1次交付失败（红叉记0，非原始物理量为0）。黑竖线为全部纳入结果的均值；$n/N$为已确定成绩／纳入数，颜色区分系统。}{Per-output congestion, timing, and quality on aligned scales and run offsets. Each panel contains 30 geometry-passing outputs (colored points), nine geometrically invalid outputs, and one delivery failure (red crosses, scored zero rather than zero raw measurements). Black ticks: all-outcome means; $n/N$: scored/included counts; colors: systems.}}
\label{fig:distribution}\end{figure}}

\newcommand{\costfigure}{%
\begin{figure}[!htbp]\centering
\includegraphics[width=\linewidth]{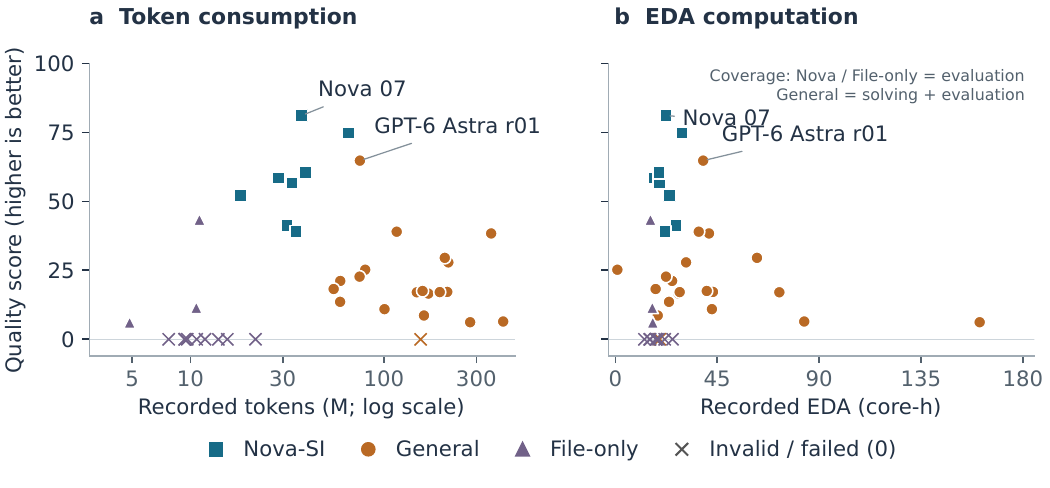}
\caption{\bi{逐结果配对的交付质量与记录资源。两个面板均含8份Nova产物及32条对照结果。(a) 总token；各点采用对应产物的token测量值。(b) EDA核时：Nova与File-only为外评，General为已记录求解与评估的日志下界。叉号保留宏几何非法和交付失败0分结果的资源消耗。每点的质量和资源来自同一研究对象；这不是等预算比较，核时也不是完整系统计算量。}{Delivered quality and recorded resources paired by outcome. Both panels contain eight Nova outputs and 32 baseline outcomes. (a) Total tokens, paired with the corresponding output. (b) EDA core-hours: external evaluation for Nova and File-only; recorded solving plus evaluation lower bounds for General. Crosses retain resource use of geometrically invalid outputs and the delivery failure, all scored zero. Quality and resources belong to the same included output at each point; this is not an equal-budget comparison or an exhaustive system-compute account.}}
\label{fig:cost}\end{figure}}

\newcommand{\novadetailfigure}{%
\begin{figure}[!htbp]\centering
\includegraphics[width=\linewidth]{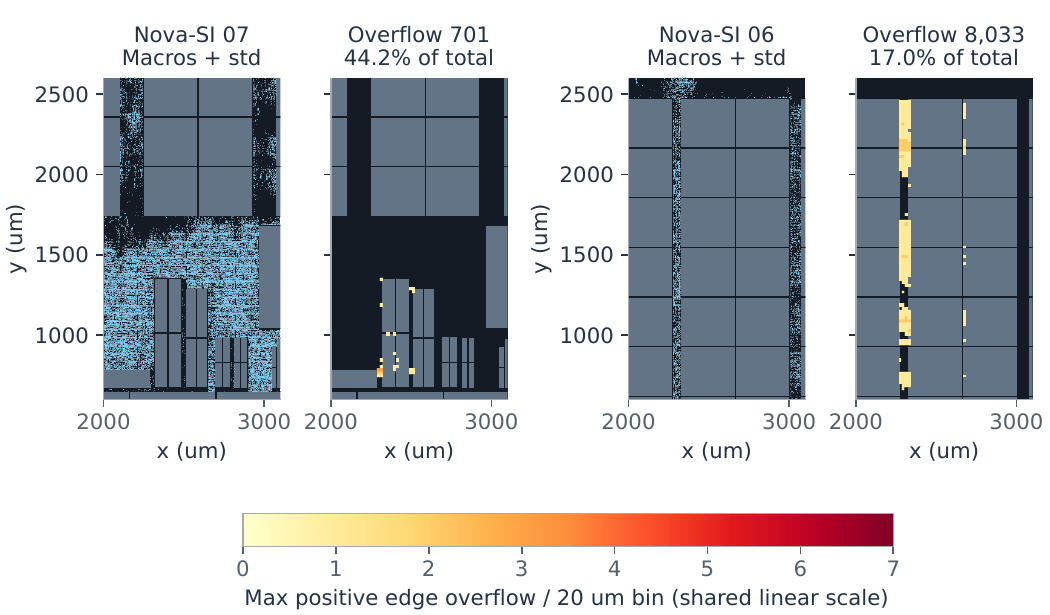}
\caption{\bi{相同窗口中的Nova高低分对照。两例均取横坐标2000--3100、纵坐标600--2600\,$\mu$m（左闭右开），窗口在观察整图后选定。每对左图为macro与实际std，右图为全部层每20\,$\mu$m显示格的最大正overflow，共用0--7线性色阶，不可与全景图按颜色直接比值。07与06窗口内正overflow边分别为525和7,869条，总overflow为701和8,033，按原生边中点统计。后者仅占06全局overflow的17.0\%，不足以解释整体分差。}{A common spatial window: x from 2000 to 3100 and y from 600 to 2600\,$\mu$m (upper bounds excluded), selected after full-layout inspection. Each pair shows macros with actual std, then maximum positive edge overflow across layers per 20\,$\mu$m display bin, on a shared linear 0--7 scale; colors cannot be compared numerically with the full-layout panels. Native-edge midpoints give 525 and 7,869 positive-overflow edges, with totals of 701 and 8,033, for 07 and 06. The latter is only 17.0\% of 06's global overflow, so this window alone cannot explain the overall score difference.}}
\label{fig:nova-local}\end{figure}}

\newcommand{\timingcasetable}{%
\begin{table}[!htbp]\centering\papertablestyle
\caption{\bi{Nova 06与07的时序严重程度和违例广度。}{Timing severity and breadth for Nova 06 and 07.}}
\label{tab:timing-case}
\begin{tabular*}{\linewidth}{@{\extracolsep{\fill}}lrrrrrr@{}}\toprule
&\multicolumn{2}{c}{\tgroup{WNS (ns)}}&\multicolumn{2}{c}{\tgroup{TNS (ns)}}&\multicolumn{2}{c}{\tgroup{\bi{违例数}{Violations}}}\\
\cmidrule(lr){2-3}\cmidrule(lr){4-5}\cmidrule(lr){6-7}
\bi{路径族}{Path family}&06&07&06&07&06&07\\\midrule
reg-to-reg&$-2.041$&$-0.231$&$-54.420$&$-18.772$&467&429\\
reg-to-mem&$-2.491$&$-0.575$&$-236.917$&$-4.106$&375&33\\
mem-to-reg&$-0.998$&$-0.309$&$-20.041$&$-1.626$&82&18\\\bottomrule
\end{tabular*}
\tablenote{\bi{reg为寄存器，mem为存储器。WNS为最坏负裕量，TNS为总负裕量；每个量分别取四场景最坏值；不把不同场景的违例相加。WNS和TNS越接近0越好，违例越少越好。两份不同产物，不是匹配路径的优化前后测量。}{Here reg means register and mem means memory; WNS is worst negative slack and TNS total negative slack. Each quantity takes its own worst value across four scenarios; violations are not summed across scenarios. WNS and TNS closer to zero, and fewer violations, are better. These are two different outputs, not before/after measurements of matched paths.}}
\end{table}}

\newcommand{\feedbackordertable}{%
\begin{table}[!htbp]\centering\papertablestyle
\caption{\bi{19条已测General运行的PG+placement尝试顺序分类。}{PG+placement attempt ordering in the 19 measured General runs.}}
\label{tab:feedback-order}
\begin{tabular*}{\linewidth}{@{\extracolsep{\fill}}lcl@{}}\toprule
\bi{首次／选定候选}{First / selected candidate}&$n$&\bi{记录中的执行顺序}{Recorded execution order}\\\midrule
\bi{相同}{Same}&4&\bi{仅一次尝试}{One attempt only}\\
\bi{相同}{Same}&10&\bi{首次完整观测之前有尝试}{Attempts before first complete observation}\\
\bi{相同}{Same}&2&\bi{首试完整，之后仍有尝试}{First attempt complete; later attempts follow}\\
\bi{不同}{Different}&3&\bi{选定另一个候选}{Another candidate selected}\\\bottomrule
\end{tabular*}
\tablenote{\bi{类别互斥；第2类可同时含后续尝试。计数描述执行顺序，不等于不同布局数，也不区分失败、修复与回退。}{Classes are mutually exclusive; the second may also include later attempts. Counts describe execution order, not distinct layouts or a classification of failures, repairs, and rollbacks.}}
\end{table}}

\author{%
  \raisebox{-0.25\height}{\includegraphics[height=22pt]{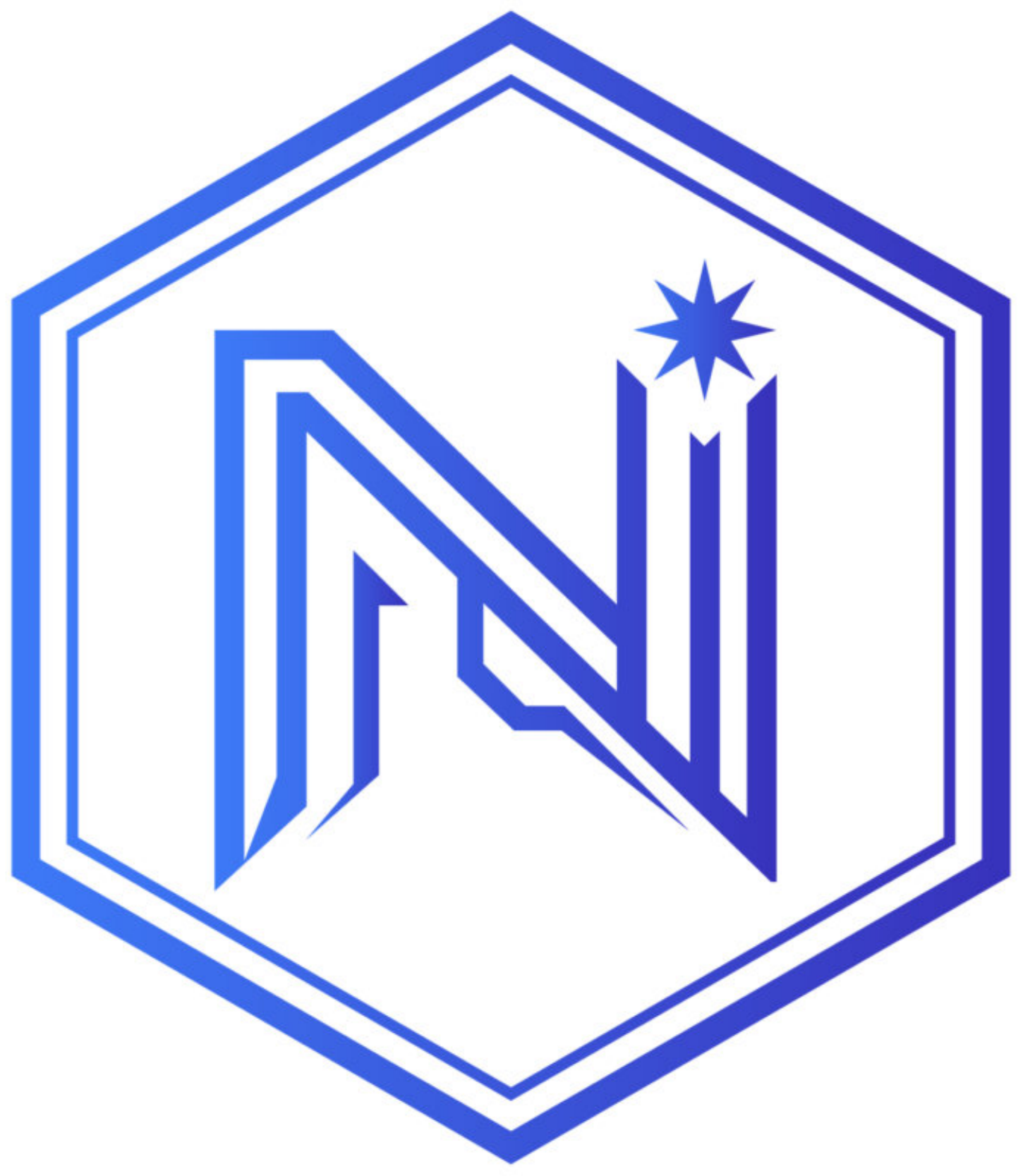}}%
  \hspace{0.6em}{\large\bfseries Nova-Silicon Team}\\[5pt]
  {\normalfont\small\hypersetup{hidelinks}%
   \href{mailto:xinxingyuan-sh@nova-silicon.com}{\texttt{xinxingyuan-sh@nova-silicon.com}}}%
}
\hypersetup{pdfauthor={Nova-Silicon Team}}

\begin{document}
\title{Evaluating AI Agents for Industrial-Level Chip Floorplanning:\\A Framework and Empirical Study}
\maketitle

\begin{abstract}
Floorplanning is an early step in turning a logical circuit into a physical chip: it positions large circuit modules, called hard macros, and constrains the space available for subsequent logic placement and wiring. These decisions affect congestion and timing, but their physical consequences emerge through downstream implementation. For AI agents, the task requires translating engineering constraints into spatial decisions whose quality cannot be judged by successful tool execution alone. We present an evaluation framework that connects submitted layouts to a prescribed downstream implementation flow, checks geometric validity, measures physical quality, and reports resource use. On one industrial design whose 99 macros occupy approximately 71\% of a fixed L-shaped layout region, we compare our specialized system Nova-SI with general-purpose engineering agents and agents restricted to file access across 40 evaluated outcomes. Under our 0--100 composite quality score, Nova-SI achieves a mean of 57.97 without online feedback from the prescribed power-and-ground network construction and placement evaluation, compared with 31.17 for the strongest measured general-purpose configuration, GPT-6 Astra. The difference is supported by raw congestion and timing measurements and persists across scoring-sensitivity checks. Layout-level analysis further shows that similar global utilization can conceal substantially different local congestion. Together, the protocol and case study assess agents through the physical consequences of their delivered layouts and provide evidence about system performance on this industrial instance.
\end{abstract}

\section{Introduction}\label{sec:intro}
Tool-using AI agents are being evaluated on increasingly substantial engineering tasks, including machine-learning experiments and command-line workflows~\citep{chan2024mle,merrill2026terminal}. Chip design presents a further question: can an agent turn its actions into a physically useful engineering artifact? A successful script or plausible explanation is not sufficient. The delivered design must preserve the intended circuit and constraints, and its quality may become apparent only after downstream tools have implemented decisions made earlier.

We study floorplanning, an early stage in turning a logical circuit into a physical chip. The circuit specifies components and their connections; floorplanning assigns positions and orientations to large, fixed-shape modules called hard macros. Electronic design automation (EDA) tools then place the smaller logic elements, called standard cells (std), and construct their connections. Macro choices determine the space and passages available to those later stages. Even when every macro fits, concentrated connection demand can exceed local routing capacity, causing congestion, while long or slow paths can miss timing requirements. Early placement decisions thus constrain later implementation. Geometric packing and wirelength proxies alone do not capture this quality~\citep{wang2024chipbench,li2026placereasoner}.

Floorplanning therefore tests whether agents can interpret project files and translate engineering constraints into coupled spatial decisions. Agents express these decisions in scripts or placement constraints and, where permitted, revise them using expensive downstream feedback. EDA tools then perform std placement and optimization. Evaluating the resulting system requires specifying both what the agent controls and how its submission becomes the measured design.

Our industrial instance makes these requirements concrete: over 230,000 circuit-component instances must be implemented within a prescribed L-shaped region, with hard macros of varying sizes alone occupying approximately seven-tenths of its area. The missing corner excludes apparently available space within the bounding rectangle; the remaining area must accommodate logic and connections. Layout decisions therefore jointly organize large modules, smaller modules, and local passages rather than merely minimize a bounding box. Evaluating this instance asks not only whether an agent can produce legal coordinates, but also what physical conditions those coordinates leave for downstream implementation.

We address this evaluation problem through a framework and an industrial case study (Figure~\ref{fig:task}). The framework connects a fixed task contract, submission-to-measurement checks, and a prescribed downstream flow to a joint report of geometry, physical quality, and resources. In particular, it distinguishes changes made by normal downstream std optimization from changes to the submitted macro layout; it also keeps invalid geometry separate from poor quality and missing measurements. These choices make the comparison interpretable even when agents use different interfaces and delivery formats.

The case study evaluates our specialized system Nova-SI alongside general-purpose and file-only agents. We first compare the quality and resources of their delivered layouts, then examine whether scoring preferences explain the observed differences, and finally use spatial and execution evidence to interpret them. Nova-SI has higher mean measured quality than the strongest observed baseline, but their individual outcomes overlap. Beyond that comparison, the study shows that similar global utilization can hide different local congestion. It also shows why equal scores at the first complete measurement and the selected delivery need not mean an absence of exploration. Our contribution is an evaluation method and an artifact-level empirical analysis of agent performance on a fixed industrial design.
\taskfigure

\section{Related Work}\label{sec:related}
\paragraph{Physical design and downstream evaluation.}
Fixed-outline floorplanning studies emphasize boundary constraints, while Hier-RTLMP incorporates hierarchy, dataflow, and pin accessibility into large-scale macro placement~\citep{adya2001fixed,kahng2023hierrtlmp}. These works motivate attention to shapes, connections, and usable space beyond macro count.

ChiPBench evaluates power, performance, and area (PPA) across circuits after downstream implementation, highlighting gaps between surrogate objectives and end-to-end performance~\citep{wang2024chipbench}. PlaceReasoner-Beta combines geometric checks and physical feedback, evaluating routed PPA and design-rule violations for macro placement~\citep{li2026placereasoner}. PDAgent-Bench covers report interpretation, script generation, and executable physical-design workflows~\citep{li2026pdagent}. Our study follows the downstream-evaluation principle but examines delivery within an existing industrial project, stopping after power-and-ground network construction, placement, and global-route estimation. It adds artifact-level analysis of what geometry each interface makes observable, how a submission reaches its measured state, and why execution success, geometric validity, and physical quality can diverge. The emphasis is on the connection between agent actions and post-placement outcomes within a single project.

\paragraph{Agent evaluation and engineering case studies.}
MLE-bench~\citep{chan2024mle}, Terminal-Bench~\citep{merrill2026terminal}, and Long-Horizon Analog Design Bench~\citep{analog2026long} evaluate agents through tasks, tool environments, and inspectable deliverables. Complementary single-project studies investigate engineering behavior in depth: Chip-Chat~\citep{blocklove2023chipchat} follows a processor design, while Vibe Tapeout~\citep{yoshioka2026vibe} compares successive-approximation analog-to-digital converter design conditions and separately evaluates six tasks using commercial process design kits. We combine an explicit evaluation protocol with an industrial floorplanning case study, separating geometry, continuous quality, and resources. \citet{kapoor2024agents} argue for evaluating expenditure and development exposure alongside performance; \citet{agarwal2021precipice} examine the uncertainty of aggregate performance estimates with few runs. These concerns motivate our reporting of individual outcomes, invalidity, missingness, and descriptive sensitivity rather than a cross-task leaderboard.

\citet{li2026orchestration} discusses autonomous chip design in terms of action spaces, domain reasoning, and agent orchestration. That perspective motivates evaluating an engineering system beyond its language model. Our study complements this perspective with a protocol for assessing the physical outcomes of delivered layouts.

\section{Task and Evaluation Framework}\label{sec:task}
The framework answers three questions: what must an agent deliver, how is that artifact measured, and what do the reported quantities mean? The task contract fixes the engineering problem; submission checks connect the agent's artifact to the measured state; physical evaluation then separates layout quality from execution status. We instantiate this protocol on one design. Another design would require its own constraints, libraries, recipe, and justified scoring scales.
\subsection{Task definition}
The input project contains the netlist (circuit components and their connections), technology and cell libraries, a fixed L-shaped layout boundary, 107 fixed signal ports, timing and power constraints, and measurement recipes. The original database contains 235,727 instances. Its 99 functional hard macros occupy 10.043\,mm$^2$, or 71.1\% of the 14.123\,mm$^2$ layout region. They use 12 library references, with an approximately 121-fold ratio between the largest and smallest macro areas. Appendix~\ref{app:instance} gives the attributes, measurement denominators, and scale context from published designs.

These conditions constrain more than whether the macros fit. Hard macros cannot be arbitrarily resized, the fixed boundary cannot be enlarged, and the missing corner is not usable whitespace. Positions and orientations determine whether the remaining space forms connected regions or fragmented strips, while shaping connections between macro pins and fixed ports. The agent must therefore place large modules legally while reserving usable space for downstream std placement and wiring. Sufficient area does not guarantee that local connection demand can be accommodated.

Systems may choose macro coordinates and legal orientations, together with placement constraints supported by their interface. They may not delete logic, relax timing constraints specified in Synopsys Design Constraints (SDC) files, move fixed ports, or enlarge the boundary. Engineer-reference coordinates and reference groupings or group shapes are not supplied as answers. Hard macros retain fixed shapes; the downstream flow optimizes the std logic. Here, L-shaped refers to the database layout region, without an additional claim about the manufactured outline of the complete chip.

The deliverable is a layout, not an agent-written quality report. Native engineering agents provide reopenable project states; file-only agents provide declarative macro placements that an external executor applies to the design database without searching for better coordinates. The evaluated state must remain associated with that submission. Agent access to tools and feedback is an experimental condition, described in Section~\ref{sec:experiments}, rather than a property inferred from the final score.

\subsection{From submission to physical measurement}
The prescribed PG+placement evaluation builds the power-and-ground (PG) network, performs placement, and measures congestion using global routing, which estimates wire demand against available routing capacity. It also measures setup timing (whether data arrives by the required time under setup constraints), wirelength, and std area. Evaluation stops before clock-tree synthesis (CTS), detailed routing, and final signoff verification. Existing native measurements are reused when they follow the prescribed flow and can be linked to the submitted layout; otherwise, external evaluation supplies the measurement without feedback to the terminated agent. Measurements are read from native EDA reports rather than accepted from an agent's description of success.

This flow is not a passive scoring function: normal std placement, resizing, and buffer insertion change the design downstream of the macro decisions. We therefore check submission binding at the appropriate level. For 11 baselines whose submitted and measured std states differ, the audit covers all 99 macro geometries, 107 fixed signal ports, original cell names and references, and the supplied recipe. Across evaluation sources, 25 core PG, placement, and constraint files are byte-identical. These checks support interpreting the measurements as downstream outcomes of the submitted layouts, without requiring unchanged std geometry or claiming complete electrical equivalence (Appendix~\ref{app:verification}).

A run is one agent execution, which may produce several candidate layouts and invoke the evaluation flow more than once. We call a complete physical measurement an observation. The system's delivered layout determines the selected output; evaluation associates it with the latest complete matching measurement, or measures it externally when needed. Selection is therefore based on the submission, not its maximum historical score. For feedback analysis, the first observation is the earliest verifiable complete PG+placement measurement, possibly after failed attempts. Candidate identifiers, report paths, and execution order distinguish these endpoints from the full search history.

\subsection{Quality and resources}\label{sec:metrics}
Our scoring rule maps congestion, setup timing, wirelength, and standard-cell utilization to component scores $c,t,l,s$ in $[0,1]$ using fixed scales, then scores each output as
\begin{equation}
Q=100\,c^{0.35}t^{0.50}l^{0.10}s^{0.05}.
\end{equation}
Congestion includes the fraction of global-routing cells (GRCs) affected by overflow, its total, and its peak. Overflow measures demand exceeding capacity. Timing includes worst negative slack (WNS), total negative slack (TNS), and violation counts for register-to-register (reg-to-reg), register-to-memory (reg-to-mem), memory-to-register (mem-to-reg), and other native path families; negative slack means a timing requirement is missed. The utilization component is $s=1-A_{std}/(A_{die}-A_{macro})$, where the three areas describe standard cells, the layout region, and macros, respectively. Tables and figures display component scores as $C=100c$, $T=100t$, $L=100l$, and $U=100s$. Losses use fixed half-score scales independent of sample rankings or a reference layout (Appendix~\ref{app:score}). They express engineering preferences, not universal process limits.

Reported std utilization is $A_{std}/(A_{die}-A_{macro})$: its denominator already excludes macro area. It differs from macro occupancy of the whole layout and from placeable-site utilization. A std utilization near 32\% therefore does not mean that roughly 68\% of the entire design is empty, nor that local routing resources are sufficient.

Measured outputs with confirmed macro boundary or overlap violations receive zero; legal but severely congested layouts retain their low score. An agent-caused failure to complete the required delivery also receives zero. A missing report alone does not establish such a failure: terminal records must identify the outcome. All included outcomes are resolved in this study. Normalized components and total quality average all included outcomes, with invalid geometry and delivery failure scored zero; raw physical means use only measured, geometry-passing layouts. A zero quality score is thus not a zero-congestion measurement.

The multiplicative quality score limits compensation: good wirelength alone cannot cancel arbitrarily poor timing or congestion. We assess sensitivity to the weights and scales to characterize how these preferences affect the comparison. Resources comprise input/output tokens, EDA CPU-core hours, and PG+placement attempts, including recorded failed attempts and subordinate-thread usage. Nova-SI and File-only core-hours cover external evaluation; General includes solving and evaluation activity in the available logs. Incomplete log coverage makes reported core-hours lower bounds on actual consumption. Tokens are not equal computational units across models. We do not combine quality and resources using an uncalibrated exchange rate.

\section{Results}\label{sec:results}
\subsection{Experimental setup}\label{sec:experiments}
Nova-SI (Super Intelligence) is our independently developed, locally deployed system. It contributes eight evaluated outputs, each weighted equally, without online feedback from the prescribed PG+placement flow. General has five model configurations and File-only three, each with four included runs. General can query native design databases and invoke EDA and PG+placement feedback. File-only can read and edit files and report status, but cannot run a shell, scripts, or EDA; its layout is evaluated after submission. Common project inputs thus do not imply identical observability: a binary database can be directly queryable in one interface and inaccessible to the file reader in another. Table~\ref{tab:quality} identifies model--harness configurations; reasoning settings and additional details appear in Appendix~\ref{app:setting}.

We compare deployed system configurations, not isolated models under a common time or token budget. Nova-SI versus General is the principal engineering-system comparison; File-only additionally examines delivery under a restricted interface. The study includes 40 outputs and runs: eight Nova-SI outputs and 32 baseline runs. All eight Nova-SI outputs and 31 baseline runs have complete selected-layout measurements. Of these 39 layouts, 30 pass macro-boundary and overlap checks and nine do not. One General/DeepSeek run ended with a PG-state delivery after placement timed out; this agent-caused delivery failure scores zero, without imputing physical measurements. Thus all 40 included outcomes contribute to quality and resource summaries.

\subsection{Layout quality}
\qualitytable
Nova-SI's mean measured quality is 57.97, versus 31.17 for General/GPT-6 Astra, the strongest observed baseline (Table~\ref{tab:quality}). Their congestion components are 34.67 versus 13.87, and timing components 79.68 versus 59.45. Section~\ref{sec:nova-astra} examines the physical quantities and individual outcomes behind this difference; Appendix~\ref{app:timing} gives the raw tables.

Aggregate differences do not imply uniform dominance. Nova-SI spans 39.01--81.09 and General/GPT-6 Astra 17.19--64.75; one General output exceeds several Nova-SI outputs (Figure~\ref{fig:distribution}). Nine of 12 File-only outputs have invalid macro geometry: all four DeepSeek-V4.1-Flash-1017 and all four GLM-5.3-Flash outputs score zero, while GLM-5.3 has one invalid output and a configuration mean of 15.25. Favorable raw overflow cannot offset a boundary violation. Two audited File-only failures used assumed geometry, with submitted macro poses preserved in measurement (Appendix~\ref{app:cases}). General/DeepSeek-V4.1-Flash-1017 has three measured layouts and one terminal delivery failure, giving a mean quality of 16.18 over all four runs. General/GPT-6.1 Sol and GLM-5.3 have four measured layouts each, with means of 22.73 and 18.92. Appendix~\ref{app:observations} separates physical-measurement coverage from zero-scored delivery failures.
The historical engineer floorplan scores 91.88, above Nova-SI's selected-output maximum of 81.09 and the strongest baseline maximum of 64.75. This is an earlier placement reference, not signoff PPA or a matched rerun; unavailable costs are left blank. Maxima describe collected outcomes, not equal-budget best-of-$k$ estimates.
\distributionfigure

\subsection{Recorded resources}\label{sec:cost}
Token use across the eight Nova-SI outputs ranges from 18.10 to 65.54 million with a mean of 36.13 million; recorded EDA use averages 22.49 core-hours. Resource means in the main table retain all included outcomes. Figure~\ref{fig:cost} pairs all 40 quality scores with their own resources, retaining geometrically invalid outputs and the delivery failure at zero quality with their expenditure intact. File-only agents have no online EDA access but still incur external measurement costs.

The plotted core-hours cover different activity, and token records may include shared upstream work; the scatter is not a complete-system Pareto frontier. Appendix~\ref{app:cost} gives input/output splits, PG+placement attempts, and per-output records.
\costfigure

\FloatBarrier
\section{Analysis}\label{sec:analysis}
The aggregate results raise three questions: which physical quantities account for the system gap, why can apparently similar layouts differ so much, and how do execution records explain the delivered layout? We address them using configuration-level measurements, matched spatial views, and existing execution records. These are explanations at different levels of evidence, not interchangeable causal tests.

\subsection{Nova-SI versus the strongest observed baseline}\label{sec:nova-astra}
The Nova-SI--General/GPT-6 Astra comparison separates physical quality from basic deliverability: all eight Nova-SI outputs and all four Astra outputs pass the macro-boundary and overlap checks and have complete measurements. Their 26.80-point mean-quality gap therefore does not arise from zeroing invalid layouts or omitting unmeasured runs within these two configurations. The difference also extends beyond the means: median quality is 57.52 versus 21.37, and three of Astra's four outputs score below Nova-SI's lowest output, 39.01. Astra r01 nevertheless scores 64.75, above six Nova-SI outputs. The observed advantage is a higher center and fewer very low-quality outputs in this collection, not a claim that every Nova-SI layout is better.

The congestion difference is not captured by global packing density. Mean std utilization is similar (32.27\% for Nova-SI and 33.56\% for Astra), whereas mean total overflow is 17,563 versus 281,748 and affected-GRC percentage is 0.0745\% versus 0.9586\%. Mean wirelength is also lower, at 21.89 versus 28.88\,m.

Timing provides a second, complementary comparison. For register-to-register paths, mean worst-scenario WNS is $-1.225$\,ns for Nova-SI versus $-2.352$\,ns for Astra, and TNS is $-44.773$ versus $-530.426$\,ns. Both quantities are closer to zero for Nova-SI, indicating less severe worst-path lateness and a smaller accumulated timing deficit. Register-to-memory and memory-to-register families show the same direction (Table~\ref{tab:timing}). These measurements explain which physical quantities accompany the higher component scores; they do not isolate the cause of the system difference.

Nova-SI achieves these outcomes without using the prescribed evaluation flow for online feedback; Astra's interface permits this use. Access to downstream feedback alone therefore does not explain the quality ranking. Figures~\ref{fig:nova-cases} and~\ref{fig:astra-overview} show each configuration's highest- and lowest-scoring selected layouts at the same scale. In both configurations, substantial whitespace coexists with localized congestion; subsequent close-ups examine these spatial differences. They help interpret physical failure modes, but do not identify which Nova-SI algorithmic choice causes the cross-system gap. Models, engineering priors, harnesses, and budgets differ together, so the evidence supports the evaluated system configuration rather than an isolated model or equal-budget advantage.
\novafigure
\astraoverviewfigure

\subsection{Sensitivity to scoring choices}
Removing Nova-SI's best output leaves a mean of 54.67, so one high-scoring output cannot explain the observed difference. Holding weights fixed, we multiply each of seven half-score scales by 0.5, 1, or 2. Across all 2,187 combinations, Nova-SI's measured mean exceeds the strongest baseline mean by 22.30--30.38 points. This relationship persists, but three complete orderings occur among configurations. Fourteen separate single-deletion checks with remaining weights renormalized also preserve the principal relationship (Appendix~\ref{app:sensitivity}).

The sensitivity results apply to the collected outputs and tested scale ranges. Among the 30 geometry-passing outputs, Spearman correlation is 0.980 between affected-GRC fraction and total overflow, and 0.976 between TNS and violation-count summaries. These submetrics describe strongly covarying aspects of physical quality. Sensitivity checks probe scoring preferences; they do not correct selection, shared provenance, or missingness, or establish statistical significance.

\subsection{Similar utilization, different congestion}\label{sec:nova-cases}
Macros occupy approximately seven-tenths of this layout; std utilization describes std occupancy only within the remaining geometric area. It records neither how that space connects nor which connections must pass through it. The following comparison tests whether global summaries adequately describe the physical outcomes.

Nova-SI's highest- and lowest-scoring outputs, 07 and 06, score 81.09 and 39.01 despite similar global std utilization (32.16\% and 32.48\%) and relatively close wirelengths (21.41 and 22.71\,m). Total overflow is 1,587 versus 47,124 and affected-GRC fraction 0.012673\% versus 0.179684\%. They are separate outputs, not successive optimization steps. Figure~\ref{fig:nova-cases} shows their full triptychs; Table~\ref{tab:nova-cases} lists all eight outputs.

\begin{samepage}
Figure~\ref{fig:nova-local} aligns a common local window. Output 07 distributes std through a broad lower region and connecting passages; in 06, std mostly occupies strips between tall macro columns. Macro-excluded geometric area in the window is 1.060 versus 0.372\,mm$^2$, while std area is 0.455 versus 0.077\,mm$^2$. The more congested output 06 thus has lower average local std occupancy: 20.82\%, versus 42.97\% in 07. This geometric ratio is not placeable-site utilization, but does not support an explanation based simply on more std or higher mean local density.
\par\end{samepage}
\novadetailfigure

\begin{samepage}
The window contains 525 versus 7,869 positive-overflow edges, totaling 701 versus 8,033 overflow. In output 06, 7,705 of those edges exceed capacity by only one: the local deficit is widespread rather than a few severe peaks.
\par\end{samepage}

\begin{samepage}
Fragmented space, connection demand, and routing capacity remain possible explanations; the export lists only overflowing edges, preventing a complete demand--capacity decomposition. The window covers only 17.0\% of 06's global overflow and is insufficient to explain the overall score difference. The case identifies spatial structure that global area and peak summaries omit; the available evidence does not isolate a particular geometric operation as its cause.
\par\end{samepage}

\begin{samepage}
\subsection{Timing severity and breadth are distinct}
Output 07 has better timing than 06 in all three principal families, with different patterns across metrics (Table~\ref{tab:timing-case}). Reg-to-reg worst WNS magnitude is 0.231\,ns in 07 versus 2.041\,ns in 06, whereas the maximum violation counts are relatively close: 429 and 467. For reg-to-mem, both severity and breadth differ substantially: TNS magnitudes are 4.106 and 236.917\,ns, with 33 and 375 violations, respectively. Thus, a similar count can conceal much milder violations, and a single worst path does not describe the whole timing distribution.
\par\end{samepage}
\timingcasetable

Existing path-endpoint exports also show that the worst reg-to-reg paths have different endpoints: output 06 links memory-test status logic to a controller counter, whereas 07's is internal to controller pointer logic, under the same clock and scenario. The comparison therefore describes differences in family-wise worst timing rather than a delay change along a matched path. The available detailed path report covers only input/output (IO) paths and lacks cell/net delay breakdowns for these scored paths, leaving the specific delay sources unresolved.

\subsection{Feedback and layout selection}
Among 19 measured General runs, 16 first/selected pairs reference the same candidate. Four of these have a single recorded PG+placement attempt; ten have earlier attempts before their first complete measurement, and two have later attempts. Endpoint identity therefore does not mean that no exploration occurred. Attempt order alone does not distinguish failed searches, repairs, or rollbacks (Appendix~\ref{app:observations}).

The three changed pairs illustrate different outcomes. GPT-6.1 Sol r03 declines from 43.37 to 29.52 after a reported correction to preserve the original placement-site definition and remeasurement, not simply a choice of the worse scalar score. GLM-5.3-Flash r04 uses congestion feedback to widen a notch corridor, increasing quality from 6.36 to 10.93; its report shows improved reg-to-reg timing but worse reg-to-mem timing in the same functional scenario. GLM-5.3 r04 changes from 22.25 to 22.71 despite a small increase in total overflow, from 135,768 to 136,296. A composite score need not track a single congestion quantity. Feedback can reveal tradeoffs, and the selected state need not maximize our evaluation score. These observations do not isolate the causal benefit of feedback.

\FloatBarrier
\section{Discussion and Conclusion}\label{sec:discussion}\label{sec:conclusion}
This study evaluates engineering agents through the layouts they deliver and the physical consequences of those layouts under a specified downstream flow. Its central methodological distinction is between a submission, its measured implementation, and the interpretation of that measurement. Checking this connection matters when EDA legitimately changes std geometry, when an agent cannot observe the native input geometry, and when several tool attempts precede the selected result.

The empirical comparison favors Nova-SI's mean delivered quality over the observed baselines on this instance, principally through congestion and timing, although their per-output score ranges overlap. The case analyses yield practical lessons for evaluation: inspect spatial demand rather than infer it from global utilization; read timing severity together with violation counts; and use execution records before interpreting first/selected changes as search progress. Two audited File-only failures also connect reported geometry-access problems to assumed placements and verified geometric violations (Appendix~\ref{app:cases}). These findings make the evaluation informative beyond a scalar ranking.

The fixed L-shaped boundary, heterogeneous hard macros, and downstream logic together define a concrete spatial-planning problem. Within this fixed project, geometrically feasible deliveries have substantially different physical outcomes. Linking submissions, measurements, and local layout evidence makes those differences amenable to detailed analysis. Without a rectangular-outline control, the score differences cannot be attributed to the L-shape alone.

The design was available during development; the outputs are a retrospective collection rather than independently repeated runs under a prospectively fixed experimental design. Models, engineering priors, permissions, and harnesses differ jointly. This case study therefore does not establish unseen-design generalization, isolated model contributions, or equal-budget efficiency.

Post-placement quality is not signoff or electrical equivalence: PG connectivity and design-rule checking (DRC) remain relevant beyond the macro-geometry checks used to gate scores. Scales express preferences, absolute slack is not period-normalized, and resource coverage is incomplete. Private assets, code, databases, and traces prevent complete independent replication; the protocol, scoring rules, and per-output summaries reported in this paper support scrutiny within that limit. The contribution is a specified evaluation protocol supported by a detailed industrial case study. Broader designs and controlled system comparisons would be needed to test how far these observed differences extend.

\bibliographystyle{plainnat}
\begingroup\small\raggedright\setlength{\bibsep}{1.5pt}
\bibliography{references}
\endgroup
\clearpage\appendix
\section{Contributions}

\par
\noindent
\begin{minipage}{\textwidth}

\subsection{Project Lead}
Linyang~Li

\subsection{Core Contributors}
Yue~Shen, Yongkang~Chen

\subsection{Contributors}
Jianxi~Xue, Zhishu~He, Hanlin~Chen, Jianqiu~Chen, Jianze~Liang, 
Le~Gao, 
Peiji~Li,
Huanlong~Zeng, Tianyi~Lyu
\end{minipage}
\par

\section{Full Scoring Definition}\label{app:score}
A nonnegative loss $z$ is converted using a half-score scale $h$. Here $p$ is the fraction of GRCs with overflow, $O$ is total overflow, $M$ is peak overflow, and $\ell$ is reported wirelength in meters.
\scoreequations
For timing family $j$, each scenario first takes the worst negative WNS magnitude and separately sums negative TNS magnitudes and violation counts. Each quantity is then maximized across four scenarios to obtain $W_j,N_j,V_j$. The maxima can come from different scenarios; they are not deduplicated cross-scenario violation totals. The scenarios combine functional/scan modes with 125$^\circ$C/$-40^\circ$C under ssgnp, 1.08\,V, cworst settings. The prescribed timing mode is graph-based analysis (GBA).

The $rr,rm,mr$ families correspond to REG2REG, REG2MEM, and MEM2REG; other combines REG2GATE and default. Native INPUT, OUTPUT, and INOUT groups are excluded. The default group retains the tool's native classification; no path-by-path endpoint reclassification is assumed. Ungated component scores are $C=100c,T=100t,L=100l,U=100s$; $U$ is the utilization score, not the raw utilization percentage.

For the first complete measurement of General/GLM-5.3 r04, the report lists residual paths under automatic clock-named groups rather than a literal default group. These residual groups are mapped to the default bucket after verifying their membership and confirming zero TNS and violation counts for every component in the original report. Named reg/mem families remain unchanged; no timing quantities from a reopened state replace the original measurement.

Let $N$ denote the number of included outcomes, $\mathcal O$ the subset with complete physical measurements, and $g_i=0$ for confirmed macro boundary or overlap violations and $g_i=1$ for outputs passing those geometric checks. Terminal delivery failures contribute zero. Because every included outcome is resolved, the reported mean is
\begin{equation}
\qscore=\frac{1}{N}\sum_{i\in\mathcal O}g_iQ_i,\qquad Q_{\max}=\max\bigl(\{g_iQ_i:i\in\mathcal O\}\cup\{0\}\bigr).
\end{equation}
Normalized component means in the main table apply the same $g_i$ and denominator $N$, with terminal delivery failures contributing zero. Raw physical means use only the measured $g_i=1$ subset. The failed run has no invented congestion or timing measurements. This is a delivered-quality summary, not a certificate of complete electrical validity.

A wirelength half-score scale of $180$\,m gives approximately 90 points at 20\,m. It is a fixed preference for this study, not a claim that 20\,m is satisfactory for every design. The utilization score favors lower utilization within this fixed die and logic; this preference does not imply that lower utilization is universally better or that the die can shrink.

\section{\bi{实例属性与设计规模背景}{Instance Attributes and Design Context}}\label{app:instance}
\bi{表~\ref{tab:instance}区分原始输入属性与placement后指标。39份已测产物的99个功能宏均与输入的实例名、库reference和朝向无关的尺寸一致，原生die/core边界顶点也一致。尺寸比较容差为0.00021\,$\mu$m，覆盖原生坐标导出精度；不要求各布局的宏坐标相同。}{Table~\ref{tab:instance} separates original-input attributes from post-placement metrics. Across all 39 measured outputs, the 99 functional macros match the input instance names, library references, and orientation-independent dimensions; native die/core boundary vertices also match. The dimension tolerance is 0.00021\,$\mu$m, covering native coordinate-export precision. Macro positions need not match across layouts.}

\begin{table}[!htbp]\centering\papertablestyle
\caption{\bi{固定实例的输入属性。面积按原生多边形或宏尺寸计算，非placement优化后std统计；die/core为数据库布局边界。}{Input attributes of the fixed instance. Areas come from native polygons or macro dimensions, not optimized post-placement std statistics; die/core denote database layout boundaries.}}
\label{tab:instance}
\begin{tabular*}{\linewidth}{@{\extracolsep{\fill}}lc@{}}
\toprule
\bi{属性}{Attribute} & \bi{值}{Value} \\
\midrule
\bi{原始实例总数}{Original instances} & 235,727 \\
\bi{功能非宏实例 / 功能硬宏 / physical-only}{Functional non-macros / hard macros / physical-only} & 235,618 / 99 / 10 \\
\bi{功能宏的库reference数}{Functional-macro library references} & 12 \\
\bi{固定信号ports}{Fixed signal ports} & 107 \\
\bi{布局区域 / core面积（mm$^2$）}{Layout-region / core area (mm$^2$)} & 14.123370 / 14.103612 \\
\bi{功能宏总面积（mm$^2$）}{Total functional-macro area (mm$^2$)} & 10.042939 \\
\bi{宏/布局区域面积}{Macro / layout-region area} & 71.1\% \\
\bi{宏/core面积}{Macro / core area} & 71.2\% \\
\bi{最小 / 最大宏面积（mm$^2$）}{Smallest / largest macro area (mm$^2$)} & 0.004254 / 0.514510 \\
\bi{最大/最小宏面积比}{Largest / smallest macro area} & 120.95 \\
\bottomrule
\end{tabular*}
\end{table}

\bi{公开设计为本实例的规模提供参照。ChiPBench官方数据表覆盖20个设计、10--136个宏及26,667--427,003个cells~\citep{chipbenchdata}；Hier-RTLMP的Table II覆盖25--760个宏及95K--2,529K个std~\citep{kahng2023hierrtlmp}。本例的宏数量与实例数量处于这些公开设计覆盖的范围内。不同来源的cells与std分类不能直接等同。}{Published designs provide context for the instance's scale. ChiPBench's official dataset table spans 20 designs, 10--136 macros, and 26,667--427,003 cells~\citep{chipbenchdata}; Table II of Hier-RTLMP spans 25--760 macros and 95K--2,529K standard cells~\citep{kahng2023hierrtlmp}. Its macro and instance counts lie within the ranges covered by these published designs. Their cell and std classifications are not interchangeable.}

\bi{两个维度补充了宏数量的信息。Hier-RTLMP所列实例的最大/最小宏面积比为1--18.52，本例为120.95；DG-RePlAce的六个加速器实例宏/core面积为46\%--60\%，本例为71.2\%~\citep{kahng2023hierrtlmp,kahng2024dgreplace}。这些参照使宏尺寸差异和面积占比具有具体背景，但不构成难度排名：工艺、时序要求、模块定义及评价阶段不同，本例还具有固定凹形边界。DG-RePlAce另外报告的整体Util包含2\,$\mu$m宏halo，不能与本文的std利用率混用。}{Two additional dimensions complement macro count. Hier-RTLMP lists largest-to-smallest macro-area ratios of 1--18.52, versus 120.95 here. The six accelerator instances in DG-RePlAce have macro/core occupancy of 46\%--60\%, versus 71.2\% here~\citep{kahng2023hierrtlmp,kahng2024dgreplace}. These comparisons contextualize size heterogeneity and macro occupancy, not difficulty: technologies, timing requirements, module definitions, and evaluation stages differ, and our instance also has a fixed concave boundary. DG-RePlAce's separate overall Util includes a 2\,$\mu$m macro halo and must not be conflated with our std utilization.}

\bi{宏外几何面积不是完整的可放置或可布线容量模型。std利用率以该面积为分母，未逐项扣除所有placement-site限制与halo；金属层上的布线容量还受其他约束。因此，宏面积占比、std利用率和原生拥塞测量分别描述不同性质，需要结合局部布局解读。}{Macro-excluded geometric area is not a complete model of placement or routing capacity. It is the denominator of std utilization without subtracting every placement-site restriction and halo; routing capacity on metal layers has further constraints. Macro occupancy, std utilization, and native congestion measurements therefore describe different properties and should be interpreted together with local layout evidence.}

\section{System Configurations and Measurement Coverage}\label{app:setting}
\conditionsummary
General uses gpt-6-astra and gpt-6.1-sol with a Codex harness at xhigh effort; deepseek-v4.1-flash-1017 with DeepSeek Harness; and glm-5.3 and glm-5.3-flash with a zcode harness. File-only uses the latter three models with InternBootcamp. DeepSeek and both GLM models use max effort: General/GLM requests set it explicitly, while DeepSeek and File-only inherit deployment or chat-template defaults. Effort labels do not establish a common reasoning budget.

All eight Nova-SI outputs are included with equal weight. The analysis describes these collected outputs; it does not estimate an independent-run success probability.

All 32 baseline runs have terminal outcomes. Thirty-one have complete selected-layout measurements. In General/DeepSeek-V4.1-Flash-1017 r03, placement reached its agent-configured 21,600-second time limit, after which the agent delivered a layout with a constructed PG network and terminated without completing placement. This is a delivery failure rather than an unresolved measurement: it receives zero quality and remains in all-run resource accounting. Its physical quantities remain absent.

\section{Native Measurements and Submission Binding}\label{app:verification}
We verify both numerical fidelity and the connection between submissions and measured states. Congestion totals, wirelength, and WNS, TNS, and violation counts of the three principal timing families are checked against native reports. Recomputed total and component scores for all 39 measured outputs differ by less than $2.2\times10^{-14}$. Nova-SI reports match hashes recorded during evaluation. Twenty-five core PG, placement, and constraint recipe files are byte-identical across evaluation sources; report-export files differ.

For 11 baselines with different submitted and measured std states, checks cover all 99 macro geometries, 107 fixed signal ports, the original 235,727 cell names and references in the submitted state, and 27 recipe files against the provided native recipe. Source-to-PG-to-placement records connect submission and measurement. The submissions include unplaced std and earlier placed states. One retained-report recovery was checked against 343,865 exported objects, with no layout or timing reoptimization. These are normal stage differences, not a requirement to submit the final post-placement database or a formal equivalence claim for all electrical states.

\section{Raw Physical Quantities and Timing}\label{app:timing}
\FloatBarrier
\suppressfloats[t]
\physicaltable
The engineer reference is included in the physical, timing, and per-output quality tables. It has overflow 142, peak 2, affected GRC 0.001383\%, wirelength 20.315\,m, and std utilization 32.134\%. Its historical PG state and recipe provenance preclude a controlled human--agent comparison.
Nova-SI scores in output-ID order are 58.42, 56.62, 60.39, 41.33, 52.07, 39.01, 81.09, and 74.84. Every included output contributes equally.
\timingtable
\novatable
\FloatBarrier

\section{Scoring Sensitivity}\label{app:sensitivity}
Removing Nova-SI's highest-scoring output reduces its mean from 57.97 to 54.67. This checks concentration in a single outcome; it does not remove selection bias or produce a new independent sample.

We multiply each top-level quality weight by factors of 0.5, 1, and 2 and renormalize the four weights to sum to one. Nova-SI retains the highest measured mean in all 81 settings, with a gap of 22.97--27.36 points to the strongest baseline mean. Half-score scales and the collected outputs remain fixed. This tests sensitivity to the relative importance of components, not universal scale validity or statistical significance.

The seven half-score scales (affected GRC, overflow, peak, WNS, TNS, violation count, wirelength) each take factors 0.5, 1, and 2, with weights fixed: $3^7=2,187$ settings. Nova-SI's measured-mean gap to the strongest baseline, GPT-6 Astra, remains positive at 22.30--30.38. Three complete configuration orderings occur, with changes among the baseline configurations. Separately, we delete one congestion submetric, one loss shared across timing families, one timing family, or one top-level component, renormalizing remaining weights within the affected group. These 14 checks give gaps of 16.99--28.94. The two analyses are separate, not an exhaustive joint perturbation. The sensitivity analysis was not preregistered before data collection.

Across 30 geometry-passing outputs, Spearman correlations are 0.980 for GRC/overflow, 0.845 for overflow/peak, 0.786 for GRC/peak, and 0.976 for TNS/violation-count summaries. Timing summaries sum family-wise worst-scenario quantities, not deduplicated paths across scenarios. Correlation describes covariation and deletion probes score dependence; neither identifies physical causality.

\section{First and Selected Physical Observations}\label{app:observations}
\coveragebounds
Table~\ref{tab:coverage-bounds} distinguishes measured outcomes from terminal delivery failure. The latter has an assigned zero score, not an unknown score requiring imputation or a missing-outcome bound. Physical summaries and first/selected comparisons require actual measurements and therefore exclude this failed run.

\pairedtable
\feedbackordertable
The first observation is the earliest retained complete PG+placement measurement, not necessarily the first attempted execution. The selected observation is the complete measurement associated with the system's delivered layout, as defined in Section~\ref{sec:task}. For the 16 General pairs with identical candidate references, matching report paths and hashes also confirm that both endpoints reuse one measurement. Earlier attempts without complete measurements have no physical-quality score; their recorded resource use is included. Nova-SI and File-only reuse their post-submission evaluation at both endpoints by construction. The human reference has no retained first/selected pair and is excluded from Table~\ref{tab:paired}.

The changed Sol r03 selection record states that a grid candidate worsened WNS and congestion, and that the retained automatic layout was repaired for original-site preservation before remeasurement. Its final report describes restoring std orientations while preserving macro poses. This is agent-reported selection intent, not a controlled attribution of the score decline. Flash r04's report describes widening a notch corridor from 60 to 199\,$\mu$m following congestion inspection. In its functional 125$^\circ$C comparison, reg-to-reg TNS changes from $-37{,}357.2$ to $-1{,}364.4$\,ns, whereas reg-to-mem WNS worsens from $-4.103$ to $-7.678$\,ns. This reported intervention illustrates a tradeoff between timing families.

\section{Resource Definitions and Coverage}\label{app:cost}
\resourcetable
\novacosttable
$N_{\mathrm{eval}}^{\mathrm{online}}$ and $N_{\mathrm{eval}}^{\mathrm{external}}$ count recorded online and external PG+placement attempts, including failures, but need not cover every custom EDA call. Tokens include recorded root and subordinate threads. EDA core-hours sum logged CPU seconds divided by 3600, including failed calls; incomplete coverage makes them lower bounds. Wall time depends on scheduling and concurrency and cannot substitute for CPU use.

Token counts are associated with the corresponding included outcomes. Nova-SI and File-only EDA figures measure external evaluation; General includes solving and evaluation activity in the available logs. Shared upstream token attribution and incomplete log coverage prevent interpreting all rows as independently metered, complete-system totals. We retain the components rather than assigning a resource score with an unsupported token-to-CPU exchange rate.

\FloatBarrier
\section{Scope and Evidence for Spatial Case Studies}\label{app:cases}
\casefigure
General/GPT-6 Astra runs r01 and r02 have global std utilization of 33.16\% versus 33.94\%, but overflow of 4,204 versus 204,229. In Figure~\ref{fig:cases}, r02's horizontal small macros partition strips containing std; several entrances coincide with strong hotspots absent at the same intensity in r01. Geometry and optimized std properties change together, so this comparison locates a difference without isolating the causal effect of channel width.
The task example is Nova-SI's mid-scoring output 02. The main comparisons use its highest/lowest outputs 07/06 and the General/GPT-6 Astra r01/r02 pair. The single failed run without completed placement has no physical measurement panels; it remains included at zero quality in the outcome summaries.

Each triptych shares a coordinate system across macros and boundaries, actual std geometry, and global-routing congestion. Comparison panels use common scales and state their layer aggregation. All figures are reconstructed from native coordinate and measurement exports. Std geometry and congestion come from the corresponding post-placement measurement state.

Nova-SI's mid-scoring output 02 has peak overflow 12, higher than output 06's 7, but lower total overflow (10,465) and a better timing component. One peak cannot replace spatial extent. Worst-scenario reg-to-reg WNS is $-0.231$ and $-2.041$\,ns for outputs 07 and 06; the congestion maps describe spatial patterns, whereas identifying the source of this timing difference requires cell/net delay breakdowns for the corresponding paths.

\invalidfigure[!p]
Figure~\ref{fig:invalid} shows all out-of-bound objects in the invalid layout. This layout receives zero under the geometric-validity rule; its raw physical measurements serve only as diagnostic evidence. Spatial overlap with a hotspot is an observed association; without matched net/path evidence or a controlled intervention, it does not establish a unique physical cause.

For File-only/GLM-5.3-Flash r04 and GLM-5.3 r03, all 99 submitted origins and orientations match the measured macro poses within the export precision ($10^{-4}$\,$\mu$m). These outputs have 53 and six macros outside the die, respectively; the latter also has 295 overlapping macro pairs. Both agents' final reports state that binary geometry was unreadable through their available interface and describe assumptions about dimensions or boundaries. GLM-5.3 estimates macro area at 4.77\,mm$^2$, versus the native total of 10.043\,mm$^2$. The unchanged macro poses place the geometric errors in the submitted layouts rather than in subsequent execution of those coordinates. The final reports suggest that limited geometry access contributed to these errors; they do not establish whether alternative extraction strategies were available.

For GLM-5.3 r03, the native tool-response trace records a UTF-8 decoding failure when reading the initial design database, followed by failures on library files. This provides tool-level evidence for the information-access difficulty reported in that run; for Flash r04, the access explanation rests on its final report. The sequence is a failed geometry read, explicit assumptions in the submitted report, then measured geometry violations with the submitted poses intact. It supports an interface-level failure explanation for this run, not a general causal attribution for all nine invalid outputs.

\end{document}